\documentclass[times,twocolumn,final]{elsarticle}

\usepackage{medima}
\usepackage{framed,multirow}

\usepackage{amssymb}
\usepackage{amsmath}
\usepackage{latexsym}

\renewcommand{\cite}[1]{\citep{#1}}
\newcommand{\bluecline}[1]{\arrayrulecolor{blue}\cline{#1}\arrayrulecolor{black}}
\newcommand{\thickhline}{\Xhline{3.5\arrayrulewidth}}

\usepackage{graphicx}
\usepackage{caption}
\usepackage{ltablex}
\usepackage{multirow}
\usepackage[table]{xcolor}
\usepackage{makecell}

\newcommand{\figref}[1]{\figurename~\ref{#1}}
\newcommand{\tabref}[1]{\tablename~\ref{#1}}
\newcommand{\secref}[1]{Section~\ref{#1}}

\usepackage{ifthen}
\newboolean{revised}
\setboolean{revised}{false}

\journal{Medical Image Analysis}

\usepackage{hyperref}
\hypersetup{pdfauthor=author}

\begin{document}

\verso{Oluwatosin Alabi \textit{et~al.}}

\begin{frontmatter}

\title{Multitask Learning in Minimally Invasive Surgical Vision: A Review}

\author[1]{Oluwatosin {Alabi}}
\author[1]{Tom {Vercauteren}}
\author[2,3]{Miaojing {Shi}\corref{cor1}}
 \cortext[cor1]{Corresponding author. \ead{mshi@tongji.edu.cn}}

\address[1]{School of Biomedical Engineering \& Imaging Sciences, King's College London}
\address[2]{College of Electronic and Information Engineering, Tongji University}
 \address[3]{Shanghai Institute of Intelligent Science and Technology, Tongji University}

\received{}
\finalform{}
\accepted{}
\availableonline{}
\communicated{}

\begin{abstract}
Minimally invasive surgery (MIS) has revolutionized many procedures and led to reduced recovery time and risk of patient injury. However, MIS poses additional complexity and burden on surgical teams. Data-driven surgical vision algorithms are thought to be key building blocks in the development of future MIS systems
with improved autonomy.
Recent advancements in machine learning and computer vision have led to successful applications in analyzing videos obtained from MIS with the promise of alleviating challenges in MIS videos.

Surgical scene and action understanding
encompasses multiple related tasks that, when solved individually, can be memory-intensive, inefficient, and fail to capture task relationships. Multitask learning (MTL), a learning paradigm that leverages information from multiple related tasks to improve performance and aid generalization, is well-suited for
fine-grained and high-level understanding of MIS data.

This review provides a narrative overview of the current state-of-the-art MTL systems that leverage videos obtained from MIS.
Beyond listing published approaches, we discuss the benefits and limitations of these MTL systems. Moreover, this manuscript presents an analysis of the literature for various application fields of MTL in MIS, including those with large models, highlighting notable trends, new directions of research, and developments. 

\end{abstract}

\end{frontmatter}

\section{Introduction}  \label{sec::introduction}

Minimally invasive surgeries (MIS)
have become increasingly popular due to their benefits, such as reduced blood loss, less pain, faster recovery times, and fewer post-surgical complications \citep{jaffray2005minimally}. However, MIS utilizes indirect vision with a limited field of view from the endoscope, making it challenging for surgeons to interpret events on the surgical scene accurately  \citep{real_time_seg_adverserial}. To overcome this, computer-aided intervention techniques that augment the information obtained through minimally invasive cameras have been proposed ~\citep{vecauteren_cai_4_cai,cv_in_surgery}.

Deep learning has demonstrated remarkable success in providing solutions to computer vision tasks for MIS, such as instrument classification \citep{wang2017deep,mishra2017learning,al2018monitoring}, instrument segmentation \citep{garcia2017toolnet,milletari2018cfcm,pakhomov2020searching,islam2019real}, and surgical scene depth estimation \citep{luo2019details,xiao2020depth,wei2022stereo,luo2022unsupervised}. Conventionally, separate models would be trained for these related tasks, which can be computationally impractical and inefficient. Therefore, there is a need for deep learning approaches that leverage information from different tasks to improve both performance and efficiency.

Multitask learning (MTL)
is a machine learning approach that seeks to improve generalisation performance by leveraging domain-specific information from multiple related tasks, with shared parameters reducing overfitting,  memory footprint and improving regularisation \citep{rich_caruana}.
MTL has been successfully applied in various fields such as natural language processing \citep{unified_mtl_nlp}, computer vision \citep{fastrcnn}, and robotics \citep{mtopt}.
In the context of MIS, MTL has primarily been utilized to enhance scene understanding. By simultaneously solving multiple tasks and leveraging knowledge from all of them, MTL offers an efficient approach for solving multiple tasks in MIS.
To ensure the widespread adoption of computer-aided intervention solutions for MIS, it is crucial to develop systems that can comprehensively understand the surgical scene, rather than addressing individual tasks separately.

In this reveiw, we aim to provide a broad
survey of multitask learning (MTL) for surgical scene understanding in minimally invasive surgeries (MIS). Our goal is for this manuscript to serve as a foundational reference for researchers and practitioners working at the intersection of multitask
learning and surgical scene understanding.

We highlight a significant methodological gap between the use of multitask learning in the computer vision and MIS communities, noting that MIS research predominantly relies on basic MTL techniques such as hard parameter sharing and linear scalarization of the loss functions.
In contrast, computer vision research has explored a wider array of MTL techniques, such as adaptive task balancing and optimization methods, flexible parameter sharing strategies, and data-efficient learning approaches to handle limited labeled data. This gap presents new opportunities for improved performance in surgical scene understanding tasks through advanced techniques and opens up novel research directions to address domain-specific challenges for combining tasks and utilizing advanced MTL techniques in MIS.

The remainder of this survey paper is divided into five sections. 
\secref{sec::review_methodology} outlines the review methodology employed, including the scope of the review, search criteria, selection process used to identify relevant literature on multi-task learning (MTL) in minimally invasive surgeries (MIS), and a brief comparison to similar review papers. 
\secref{sec::multitask_learning_natural_images} provides a quick introduction to popular deep MTL techniques commonly employed in MTL research for vision, emphasizing their key features and concepts. \secref{sec::mtl_surgical} reviews how MTL has been applied to solve multiple tasks for surgical scene understanding, examining each work extensively and identifying the current trends in various research areas. In \secref{sec::datasets} `Public datasets for MTL in MIS', presents an overview of publicly available datasets that can be used to advance MTL research in MIS.\secref{sec::discussions_and_conclusions} presents our insights on the current state of MTL research in MIS, discusses potential directions for future work, and concludes this survey by summarising the main contributions of this paper.

\section{Review methodology and related work}\label{sec::review_methodology}

\begin{figure*}[htb!]
   \centering
  \includegraphics[width=0.9\textwidth]{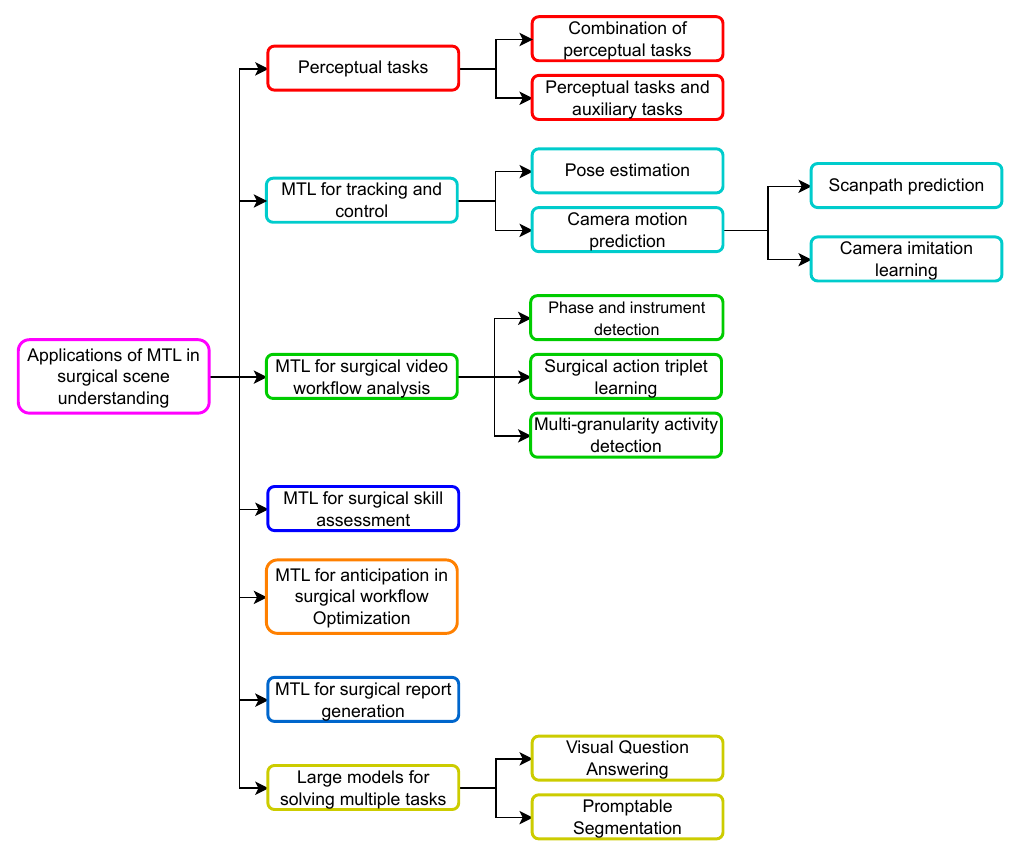}
  \caption{Overview of the application areas where multitask learning has been applied in surgical scene understanding.}  \label{fig:applications_of_mtl_in_scene_tree_diagram}
\end{figure*}

\subsection{Scope of the review} 
This review focuses on the application of multi-task learning (MTL) in minimally invasive surgeries (MIS). Specifically, we limit the scope to methods that utilize videos and/or images obtained from MIS cameras to address multiple tasks, where each task yields meaningful and relevant outputs. 
Papers that primarily focus on one task, but include auxiliary tasks to guide the learning of the primary task, were also considered for inclusion.
With the growing relevance of large models in medical research, we also review studies that leverage such models to address multiple tasks within the context of MIS.
To ensure completeness and provide a comparative perspective, we discuss a few seminal multi-task learning papers from the general purpose computer vision and machine learning communities. 
These works introduce readers to deep multi-task learning techniques from the broader field, helping to highlight which techniques are employed or underutilized in surgical scene understanding. 

Our review is structured as a narrative review, rather than a systematic one. This approach allows for greater flexibility in examining the literature, offering a more expansive view than that typically used in systematic reviews.

\subsection{Search criteria }
The works reviewed were identified through an initial search on Google Scholar using keywords "multitask learning surgical vision". Following this, we manually reviewed the references within the initially identified papers to discover additional relevant works. We also conducted author-specific searches to locate further research by key contributors in the field. Finally, thematic analysis was carried out on the studies that tend to have multiple tasks solved. Searches and selections were completed by July 2023.

\subsection{Selection criteria}
In accordance with the scope of this review, we only included publications that presented methods validated on minimally invasive surgical imaging data. Hence, we excluded works that focus on robot kinematics information, and we excluded works that focused on data from open surgeries or external operating room camera setups. 
MIS typically employs a single, small camera inserted through an incision, offering a narrow field of view on the surgical site to support precision in confined spaces. In contrast, open surgeries and external camera setups involve wide-field imaging, with good natural lighting and fewer distortions. The scene showed by  open surgeries and external camera setups have limited visual and semantic similarity to those seen in MIS. These differences make the challenges and research questions in MIS distinct from those in other types of surgery, which is why they were excluded from the review.

Following the search and selection process, 47 studies were chosen for inclusion in this review.

\subsection{Analysis of selected papers}

\figref{fig:applications_of_mtl_in_scene_tree_diagram} provides an overview of the selected studies, categorized by their application areas within surgical scene understanding where multitask learning (MTL) has been applied. In perceptual tasks, 11 studies focus on extracting visual information, such as segmentation and depth estimation, and combining tasks to improve accuracy and efficiency. MTL has also been applied to tracking and control of surgical instruments and cameras, with 7 studies addressing this. Surgical video workflow analysis, including phase recognition and action triplet learning, is explored in 13 studies, while 5 studies cover anticipation tasks, such as predicting upcoming instruments, phases and time to surgery end. 2 studies apply MTL to surgical skill assessment by predicting skill-related metrics, and 2 studies focus on surgical report generation using MTL. Finally, 7 studies investigate large models using prompts to solve multiple tasks.

\subsection{Related reviews}
To the best of our knowledge, our review is the first,  to focus specifically on multitask learning (MTL) techniques within the context of minimally invasive surgeries (MIS).
While existing reviews, such as \citet{survey_simon} and \citet{zhao2023multi}, provide valuable insights into MTL applications in general computer vision and broader medical imaging, respectively, they do not offer a targeted analysis of the specific challenges and applications of MTL within the MIS domain. 

In comparison to related reviews on deep learning in MIS, such as \citet{rivas2021review} and \citet{demir2023deep}, our work delves deeply into each study that applies MTL within MIS. We highlight the multitask parameter sharing approach and optimization method employed in each work, providing a level of detail not covered by these broader reviews. While \citet{rivas2021review} and \citet{demir2023deep} offer overviews of deep learning applications in MIS, including image analysis and surgical workflow recognition, they do not focus on works that utilize MTL specifically.

\section{Common deep MTL methodologies in computer vision}\label{sec::multitask_learning_natural_images}

This section presents a short introduction to widely used deep MTL techniques in the field of computer vision research. We present a short examination of these techniques and representative papers that showcase the utilization of each technique. By presenting the foundations and applications of MTL techniques, this section aims to equip readers with a basic understanding of the subject, setting the stage for the upcoming sections that analyze methods employed for solving multiple vision tasks in the context of MIS.

\subsection{MTL concepts}
There are different ways of categorizing MTL core concepts. For our purpose, we focus on four concepts that recurrently emerge in the MTL literature for MIS: parameter sharing, optimization and task-balancing, auxiliary objectives, and data-efficient approaches.

\subsubsection{Parameter sharing and feature representation} \label{subsec::parameter_sharing}
The mainstream multitask paradigm assumes that tasks are related as the knowledge required for solving these tasks is connected. Following this assumption, features produced for the same input sample on multiple tasks which are related should also be related. 
Hence, common feature representation learnt for multiple tasks would have more informative and robust feature representations as they take into account multiple tasks~\citep{survey_zhang}.
Additionally, each task acts as a regularization for other tasks, smoothening out noise, as features which are utilized for multiple different tasks are less likely to overfit to training samples \citep{survey_zhang}.

\begin{figure}[t]
  \includegraphics[width=0.48\textwidth]{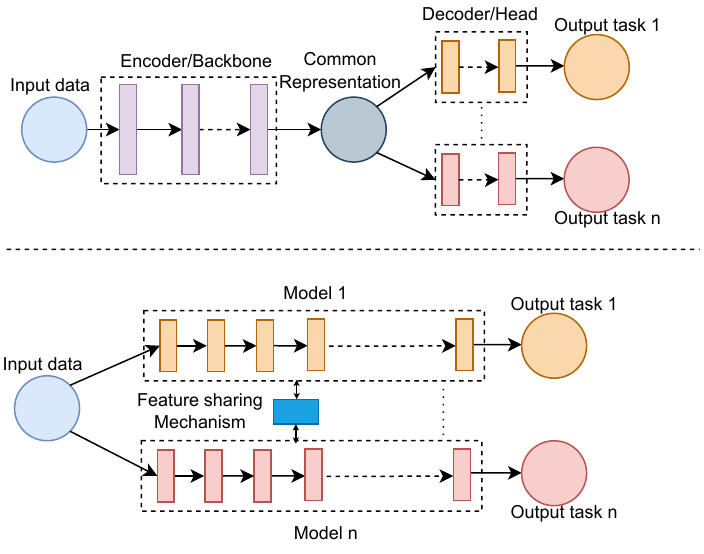}
  \caption{ Top: hard parameter sharing in deep neural networks for multitask learning, featuring a shared encoder/backbone with a common representation and separate decoders or heads. Bottom: soft parameter sharing with separate models per task and specialized feature-sharing mechanism.} 
  \label{fig:soft_and_hard_sharing}
\end{figure}

Historically, the methods for sharing feature representations in deep neural networks are classified into hard and soft parameter sharing \citep{survey_simon}. A diagrammatic representation of soft and hard parameter sharing can be seen in \figref{fig:soft_and_hard_sharing}.
Hard parameter sharing refers to architectures where tasks are to be jointly learnt utilizing the same weights and biases for some layers. These layers are aptly called \emph{shared layers}. The other layers which are not shared are called \emph{task-specific} layers. Hard parameter sharing is commonly interpreted using an encoder-decoder architecture where the encoder is shared, and the decoder is task-specific. Some network architectures for hard parameter sharing also include methods to facilitate information transfer between different decoders \citep{Xu_2018_CVPR,MTAN,Zhang_2018_ECCV}.
If $x_i$ represents input samples, $h$ denotes hidden layers, and $y_{i,t}$ represent outputs for the $i^{th}$ sample on the $t$ task, $f_{sh}(.)$ represents shared layers and $f_{task}(.)$ represents the task-specific layers, then hard parameter sharing can be summarized as  
\begin{equation}
    h_i = f_{sh}(x_{i}) \quad y_{i,t} = f_{task}(h_{i})
\end{equation}

On the other hand, soft parameter sharing does not directly share layers for each task. Instead, a separate model is used for each task. Soft parameter sharing instead shares parameters by adjusting the weights and biases of different task models based on information from other task models, leading to model weights and biases, which are functions of representations from different tasks. If $M_{t}(.)$ is a model for task $t$ and $M_t$ produces features $f_t$, then soft parameter sharing can be written as
\begin{equation}
    M_t=  R(f_{1}, f_{2} ... f_{t})   
\end{equation}
where $R$ is a feature-sharing mechanism which determines how features are shared.

A prototypical example that uses hard parameter sharing is UberNet \citep{ubernet}. UberNet is proposed as a general-purpose computer vision model for multiple tasks. The network architecture is a shared multi-resolution VGG \citep{vgg} with a branch network from each layer of the VGG for each task. The output of each task branch is fused with branches for the same branch from other layers and finally fused for all resolutions to produce predictions for each task. 
Other examples of hard parameter-sharing architectures include  \citep{Xu_2018_CVPR,multinet,multitaskcenternet,MTAN,Zhang_2018_ECCV},

Cross-stitch network \citep{Misra_2016_CVPR} is a good example of soft parameter sharing. This paper is concerned with the design decision about when to split a network into task-specific networks and general/shared parameters. To solve this, the authors propose a unified splitting architecture, which is a type of soft parameter sharing architecture. Different separate networks are connected with the cross-stitch module across various layers of the separate networks. This is done by learning a linear combination of the input activation maps from both tasks. Other examples of soft parameter sharing architectures include \citep{gao2019nddr,ruder2019latent}

\subsubsection{Optimization and task balancing} \label{subsec::optimization_and_task_balancing}
Designing networks with shared parameters and learning these parameters together raises an important question: What is the right way to optimise both tasks jointly?

\begin{figure*}[t]
  \includegraphics[width=\textwidth]{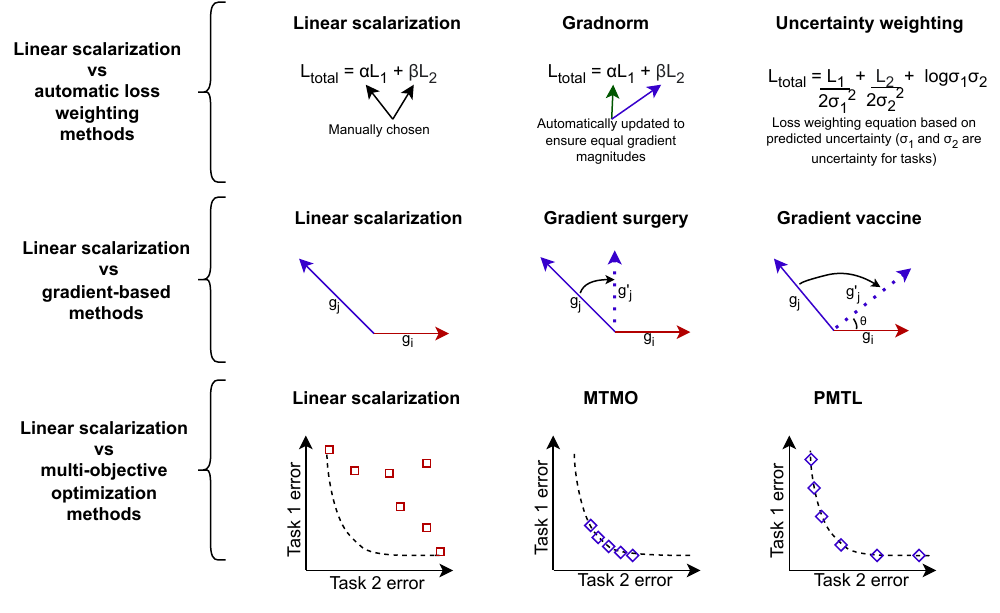}
  \caption{An overview of optimization techniques for multitask learning as discussed in \secref{subsec::optimization_and_task_balancing}.
  The classical method, linear scalarization, involves manually weighting the loss functions of all tasks.
  The first row illustrates linear scalarization against automatic loss weighting methods that dynamically adjust weights during training, such as updating weights to ensure gradient magnitude consistency \citep{gradnorm} and defining a loss weighting equation based on predicted uncertainty of each task \citep{uncertainty_weighting}. 
  The second row illustrates gradient-based approaches in comparison to linear scalarization.
  Gradient-based methods directly modify gradients to mitigate negative transfer, achieved by projecting conflicting gradients to the normal plane of the gradient of another task \citep{pcgrad_paper} or ensuring gradients are at a target angle to each other \citep{grad_vac}.
  The third row illustrates linear scalarization and multi-objective optimization techniques. MTMO \citep{MTMO_koltun} ensures that solutions are on the Pareto front, while PTML \citep{lin2019pareto} enables the selection of Pareto front solutions, favouring specific tasks. 
  \label{fig:task_balancing_and_optimization}}
\end{figure*}

\figref{fig:task_balancing_and_optimization} provides a brief overview of the various methods of optimization methods discussed in this section. \href{https://github.com/median-research-group/LibMTL}{LibMTL}\citep{lin2023libmtl} is a python repository with adaptable code implementation for the common multitask architectures and optimization strategies discussed here.

For hard parameter sharing with shared encoders and task-specific decoders, as shown in  \figref{fig:soft_and_hard_sharing}, we can write the baseline equation for the multitask loss while omitting batching considerations for simplicity, as Equation \eqref{equ::total_loss}. 
The gradient descent equation for the shared parameters can be written as Equation~\eqref{equ::linear_scalarization_shared}, and the gradient descent equation for the task-specific parameters as Equation \eqref{equ::linear_scalarization_task}:
\begin{equation} \label{equ::total_loss}
    L_{total}=\sum ^{N}_{i=0}L_{i}\left( \theta _{t},\theta _{s},X,Y\right)
\end{equation}
\begin{equation} \label{equ::linear_scalarization_shared}
    \theta _{s}=\theta_{s}-\alpha \sum ^{N\cdot }_{i=0}\dfrac{\partial L_{i}}{\partial \theta _{s}}\left( \theta _{t},\theta _{s},X,Y\right)
\end{equation}
\begin{equation} \label{equ::linear_scalarization_task}
    \theta _{t\left( a\right) }=\theta_{t\left( a\right) }-\alpha \dfrac{\partial L_{i\left( a\right)} }{\partial \theta _{t\left( a\right)} }\left( \theta _{t},X,Y\right)
\end{equation}
where $N$ is the number of tasks, $X$ is the input, $Y$ is the ground-truth, $\alpha$ is the learning rate, $L_i$ is the loss for the $i$th task, $a$ refers to the current task been considered,  $\theta _{s}$ denotes the shared parameters while $\theta _{t}$ the task specific parameters. 

From Equation \eqref{equ::linear_scalarization_shared}, we can deduce that the network weights in shared layers are affected by multiple supervision signals. Hence, the best way of balancing different task signals in the shared layers has been explored in several works \citep{MTMO_koltun,uncertainty_weighting,Dynamic_task_prioritization}.

Equation \eqref{equ::total_loss} shows the multitask loss for a network called linear scalarization - when the loss functions for $N$ tasks are simply added together. Linear scalarization is the most utilised task-balancing method \citep{hu2023revisiting}. It involves assigning a scalar weight to each task and optimising the scaled addition as can be seen in \eqref{equ::total_loss}, \eqref{equ::linear_scalarization_shared_weight}, and \eqref{equ::linear_scalarization_task_weight}  
\begin{equation}\label{equ::total_loss_weight}
    L_{total}=\sum ^{N}_{i=0}w_{i}L_{i}\left( \theta _{t},\theta _{s},X,Y\right)
\end{equation}
\begin{equation}\label{equ::linear_scalarization_shared_weight}
    \theta _{s}=\theta _{s}-\alpha \sum ^{N\cdot }_{i=0}w_{i}\dfrac{\partial L_{i}}{\partial \theta _{s}}\left( \theta _{t},\theta _{s},X,Y\right)
\end{equation}
\begin{equation}\label{equ::linear_scalarization_task_weight}
    \theta _{t\left( a\right) }=\theta _{t\left( a\right) }-\alpha w_{i\left( a\right)} \dfrac{\partial L_{i\left( a\right)} }{\partial \theta _{t\left( a\right)} }\left( \theta _{t},\theta _{s},X,Y\right)
\end{equation}
where $w_i$ is the weight assigned to a particular task, and all other notations remain the same. Despite its simplicity, linear scalarization works surprisingly well, especially when combined with techniques like grid search  \citep{xin2022current}.

Another example of a task balancing method is the automatic dynamic tuning of the weight assigned to each task at different points during training. 
\citet{uncertainty_weighting} propose a method to automatically weight losses during multitask optimization using the uncertainty in the predictions of each task. The authors reformulate the linear scalarization loss function to include a homoscedastic loss uncertainty term for each task, an uncertainty term independent of the inputs but depends only on the task problem. This method ascribes an uncertainty scalar for each task, and depending on the value of this uncertainty scalar, the gradient magnitudes of each task are weighted. Instead of using uncertainty to automatically determine loss weights, \citet{gradnorm} utilize each task's gradient values to estimate the rate of convergence of each task and corrects this rate of convergence to be equal by adjusting the weights associated with each task in linear scalarization formulation.
\citet{Dynamic_task_prioritization} introduce the notion of dynamic task prioritization to prioritize more difficult tasks. \citet{Dynamic_task_prioritization} observe that imbalance in task difficulty can result in prioritization of the easy tasks during optimization and propose to measure task difficulty as inversely proportional to the task performance. Task performance is measured by key performance metrics (KPIs) like accuracy for classification and IoU for segmentation. Dynamic task prioritization automatically prioritizes more difficult tasks by adaptively adjusting the weight of each task's loss objective based on task KPIs. 

A different task-balancing idea is the direct adjustment of the gradients for each task calculated during back-propagation instead of tuning weights to give priority to different tasks. These methods claim to be able to tackle problems with multitask optimization, such as negative interference. \citet{pcgrad_paper} propose the PCGrad method, which uses cosine similarity to check if the directions of gradients for tasks trained for multitask learning are conflicting, which can lead to negative interference. If the gradients are conflicting, the PCGrad method projects the task gradients of a task to the normal plane of the other task to remove the conflict and optimizes with this projection. Further investigation into conflicting gradients by \citet{grad_vac} reveals that the occurrence of conflicting gradients during training as measured by cosine similarity is very sparse. Hence, \citet{grad_vac} design a more proactive method for ensuring that there are no conflicting gradients called ``gradient vaccine''. \citet{grad_vac} note that the adjusted gradients using PCGrad project gradients to normals, making a cosine similarity of 0 the target in PCGrad. Gradient vaccine sets a targeted cosine similarity, which is greater than 0. By selecting a targeted cosine similarity, gradient vaccine ensures that both conflicting and non-conflicting gradients that are still very dissimilar can be projected to the target cosine similarity plane, ensuring frequent gradient update.   

Another key task-balancing idea is the interpretation of multitask optimization as multi-objective optimization, which requires a Pareto optimal solution \citep{MTMO_koltun,lin2019pareto}.
A Pareto optimal solution is reached when one objective function cannot be improved without sacrificing another objective. Pareto optimal solutions lie on a Pareto front, which is a set of optimal solutions in the space of objective functions in multi-objective optimization problems.  MTMO \citep{MTMO_koltun} and PTML \citep{lin2019pareto} are examples of papers that try to figure out the best way to optimize multiple tasks together with the Pareto principle. 
Unlike linear scalarization, which combines multiple loss functions via a system of fixed weights and optimizes the multiple losses as a single loss function, Pareto optimization is a multi-objective optimization problem.
\citet{MTMO_koltun} establishes the foundation for a Pareto optimal solution in MTL by mathematically deriving an upper bound for the multi-objective optimization problem.
This upper bound is then optimized with a multi-gradient descent algorithm. 
PTML pushes this idea further by generalizing the mathematical formulations from \citet{MTMO_koltun}. It does so by relaxing the Pareto optimal condition and allows for solutions that are as Pareto optimal as possible while still favouring certain tasks.

\subsubsection{Auxiliary objectives} \label{sec::multitask_learning_natural_images::auxiliary_tasks}
In MTL, auxiliary tasks are additional tasks that are learned simultaneously with the primary task to improve the overall performance of a model on the primary task. The idea behind auxiliary tasks is that they can provide helpful information or constraints to help the model learn a rich and robust representation of the input data, which in turn benefits the primary task \citep{liebel2018auxiliary}.
\citet{zhang2014facial} proffer a method that optimizes a main task in their application, facial landmark detection, together with auxiliary tasks like head pose estimation, facial expression recognition and age estimation. The authors propose a deep network to extract shared features with different classifier heads for each task.

\subsubsection{Data efficient approaches}
Building large-scale computer vision datasets is resource intensive \citep{Liao_2021_CVPR}. This is even more noticeable for networks that use the MTL paradigm, which requires labelling for multiple different tasks per sample. There are few publically available datasets designed specifically for working with multiple tasks \citep{zamir2018taskonomy,yu2020meta}.
Hence, problems such as limited annotations for multiple tasks and the availability of different task labels in different datasets, are often faced by researchers face when they want to utilize MTL. Consequently, there have been research works to develop MTL techniques to solve these challenges.   

A method by which multitask learning can be used to tackle issues with limited annotations for multiple tasks is with better representation learning via multitask learning. Representation learning involves learning the representation of the data that makes it easier to extract useful information for downstream tasks that require a few labelled data \citep{bengio2013representation}. Learning representations that are task-agnostic over a range of tasks would be preferred to learning single-task representations as these representations would be more robust to noise \citep{nguyen2022task}. Self-supervised MTL methods for representation learning utilize self-supervised tasks which do not require new annotations and learn representations for multiple tasks together.
These representations can then be utilized for downstream tasks with fewer amounts of data. 
An example of such work is from \citet{doersch2017multi} who utilize four self-supervised vision tasks (relative position, colourization, the exemplar task, and motion segmentation) for pre-training instead of a single pretraining task.
\citet{doersch2017multi} demonstrate that the representations learnt using their method are competitive to ImageNet pretrained representations without the need for labelling. \citet{Ghiasi_2021_ICCV} presents Multi-Task Self-Training (MuST), which is a method to harness the knowledge from specialized teacher models for different tasks to create a large multitask dataset with pseudo-labels generated from these specialized teachers. The large pseudo-label dataset is then used to train a multi-task network. The authors show that the representation generated in this multitask network generalizes well to downstream tasks. 

There are also examples in the literature for combining information in different datasets to perform MTL. \citet{li2022learning} proffer a method for combining multiple datasets collected on similar distributions but annotated for different tasks. The authors propose a method that leverages task relations between task pairs to supervise MTL task pairs jointly. They map a task pair to a supervised joint space to enable information sharing between two tasks. Then, they propose a supervised learning loss for the task with known labels and a consistency loss in the joint task space to train tasks with unknown labels.
\citet{dorent2021learning} tackles the problem of learning from domain-shifted datasets, each with single task-specific annotations. Specifically, for the brain tissue segmentation task, there are datasets with segmentation of brain structures in healthy brains, and there are datasets which segment pathologies like brain lesions and tumours, but there are no datasets with both types of labels.
The authors derive an upper bound of the loss for the joint probability problem. 

 \begin{figure}[t]
  \includegraphics[width=0.48\textwidth]{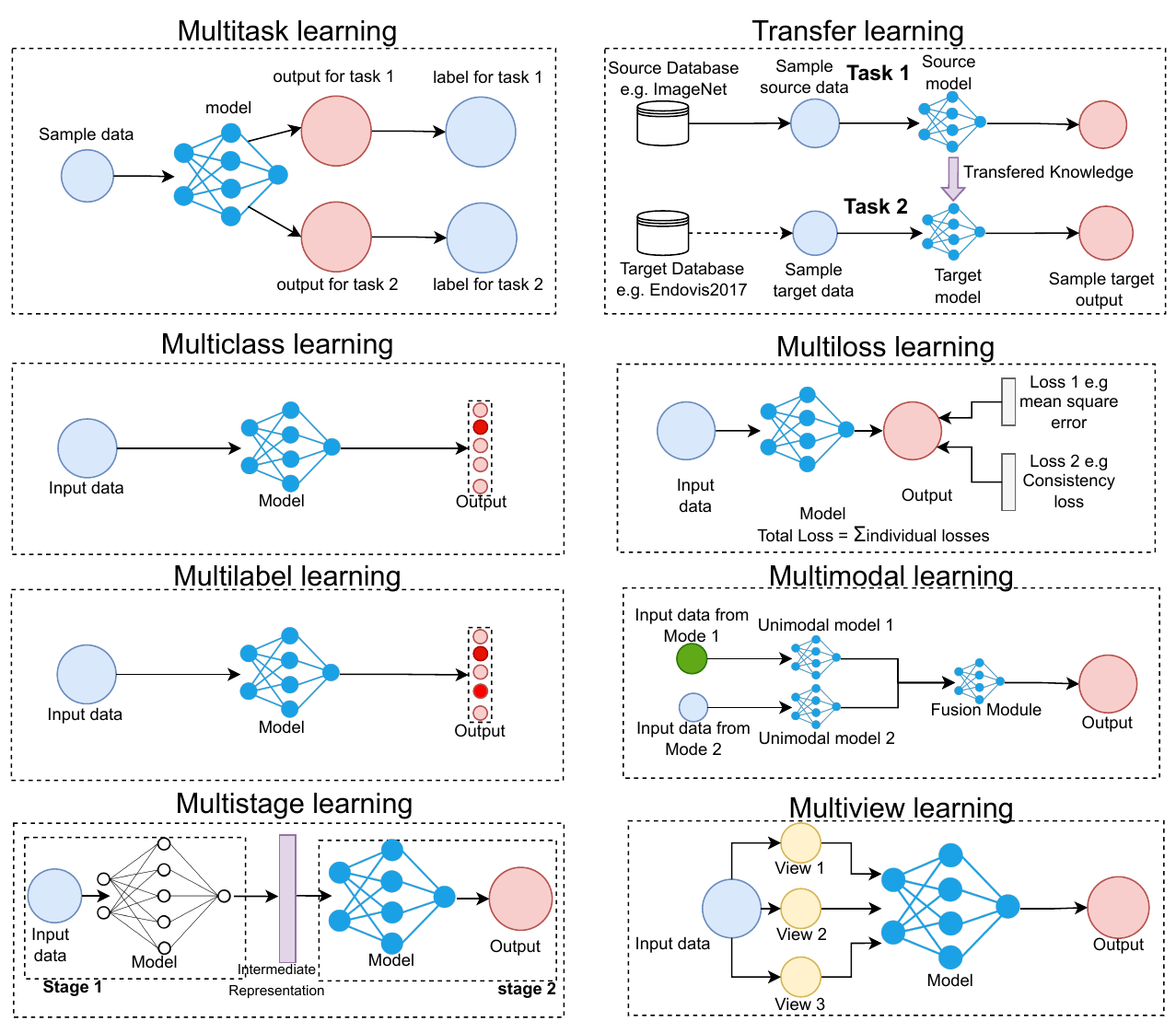}
  \caption{Diagram illustrating the key differences between multitask, transfer, multiclass, multiloss, multilabel, multimodal, multistage, and multiview learning.}
  \label{fig:mtl_and_other_learning_paradigms}
\end{figure}

\subsection{MTL and other learning paradigms}
MTL is a learning paradigm that trains a single model to perform multiple tasks simultaneously by sharing information across them, leading to better representations useful for all tasks.

The idea of using information from a different task or different outputs to improve representation is not unique to MTL and shares similarities with other learning paradigms, which can make it challenging to differentiate between them. 
Related learning paradigms include transfer learning \cite{oquab2014learning},multiview learning \cite{andrew2013deep}, multiloss learning \cite{johnson2016perceptual}, multilabel learning \cite{wang2016cnn}, multiclass learning \cite{krizhevsky2017imagenet}, multistage learning \cite{girshick2015region}, multimodal learning \cite{frome2013devise}.

A diagrammatic representation of the paradigms discussed in this section can be seen in \figref{fig:mtl_and_other_learning_paradigms}.
Despite their similarities, each paradigm has its unique features and objectives, and understanding their differences and similarities is essential for their proper application.

\emph{Transfer learning} enhances target task performance using knowledge from a related source task, while MTL trains multiple tasks concurrently.
\emph{Multi-loss learning} employs multiple loss functions for neural network training. This approach finds use in both MTL and single-task learning.
\emph{Multilabel learning} refers to problems where single data instances can have multiple class labels.
\emph{Multiclass learning} classifies input data into one of several classes, whereas MTL concurrently optimizes multiple related tasks, sharing the same input data.
\emph{Multiview learning} optimizes multiple data representations (views) with the same output.
\emph{Multistage learning} processes inputs through multiple stages. Multitask networks can have single or multiple stages based on the design.
\emph{Multimodal learning} involves learning over diverse data modalities. Multimodal learning is usually done with a separate encoder for each different modality and a fusion module. Multitask learning can have inputs of different modalities.

\section{Applications of MTL in surgical scene understanding} \label{sec::mtl_surgical}

This section delves into the applications of the multitask paradigm in the context of surgical scene understanding, where it tackles the challenge of simultaneously learning multiple subtasks to enhance both efficiency and accuracy.
Additionally, it explores the utility of auxiliary tasks to augment primary tasks in surgical scene understanding. 
In the subsequent subsections, we describe the deployment of MTL for surgical scene understanding across seven distinct categories: MTL for perceptual tasks, MTL for tracking and control, MTL for surgical workflow analysis, MTL for anticipation in surgical workflow optimization, MTL for surgical skill assessment, MTL for report generation, and large models for solving multiple tasks. 

\figref{fig:applications_of_mtl_in_scene_tree_diagram} shows an overview of the application areas where multitask learning has been applied in surgical scene understanding.

\begin{table*}[htb!]
\caption{Methods that utilize multitask learning for perceptual tasks. *HP - hard parameter sharing. } \label{perceptual_tasks:table}
\resizebox{\textwidth}{!}{
\begin{tabular}{|l|l|l|l|l|l|}
\hline
  & Publication & Tasks & Optimization & Architecture & Speed(fps)\\
\hline

\multirow{7}{*}{ \parbox{3.0cm}{Combination of perceptual tasks}}  & \citet{simultaneous_depth_tool_seg} & \thead[l]{binary tool segmentation\\unsupervised depth estimation} & linear scalarization &  shared encoder with multiple  task branches  (HP) & 172 \\
\bluecline{2-6}

 & \citet{scalable_joint_detection} & \thead[l]{tool classification\\ tool segmentation \\ tool detection} & linear scalarization &  shared encoder with multiple task branches (HP) & 22.4\\
\bluecline{2-6}

 & \citet{islam2020apmtl} & \thead[l]{tool detection\\ tool segmentation} & sequential training   &  shared encoder with multiple  task branches (HP) & 18  \\
\bluecline{2-6}

 & \citet{baby2023forks} & \thead[l]{tool classification \\ tool detection\\ tool segmentation} & sequential training  & modified Mask2former (HP) & -  \\
\bluecline{2-6}

 & \citet{zhao2022trasetr} & \thead[l]{tool classification \\ tool detection\\ tool segmentation} & linear scalarization  & modified DETR (HP) & 23 \\
\bluecline{2-6}

  & \citet{das2023multi} & \thead[l]{landmark detection\\landmark spatial relationship } & linear scalarization  &  shared encoder with multiple task branches (HP) & 10 \\
\bluecline{2-6}

 & \citet{msdesis} & \thead[l]{depth estimation \\  tool segmentation} & linear scalarization  &  shared encoder with multiple  task branches (HP) & 22 \\
\hline

\multirow{4}{*}{ \parbox{3.0cm}{Perceptual tasks and auxiliary tasks}}  & \citet{multi_contour} & \thead[l]{tool segmentation\\contour detection} & linear scalarization &  shared encoder with multiple task branches (HP) & -\\
\bluecline{2-6}

 & \citet{multi_HOG} & \thead[l]{tool segmentation \\  HoG prediction} & linear scalarization  &   shared encoder with main task branch and deep supervision auxiliary heads (HP) & -\\
\bluecline{2-6}

 & \citet{ss_depth_estimation_laparo} & \thead[l]{depth estimation\\3D left-right consistency} & linear scalarization  &  multistage network (HP) & 105 \\ 
\bluecline{2-6}

 & \citet{landmark_detection_submucosal} & \thead[l]{landmark detection\\landmark spatial relationship } & linear scalarization  &  shared encoder with multiple task branches  (HP) & 37  \\

 \hline
\end{tabular}
}
\end{table*}

\subsection{MTL for perceptual tasks} \label{section::perceptual}
In the context of this report, \emph{perceptual tasks} refer to tasks such as image segmentation, object detection, motion estimation, and depth estimation. These tasks are focused on extracting essential visual information from images or videos, with an emphasis on revealing the spatial layout, motion, semantic, and depth relationships of objects within the scene. Perceptual tasks provide valuable information about spatial layout, motion, and depth relationships in the scene. The outcomes of perceptual tasks play a critical role in providing insights into the surgical scene, serving as the foundational elements for more advanced computer vision processes, including action recognition and object tracking. By enhancing the performance of perceptual tasks, researchers can better prepare input data in a way that significantly eases and enhances the accuracy of subsequent higher-level tasks.
\tabref{perceptual_tasks:table} summarizes the papers discussed in this subsection, highlighting the multiple tasks addressed, the optimization approaches adopted, the architectural strategies used for multitask learning, and the reported speeds for each study.

\subsubsection{Combination of perceptual tasks}
MTL for perceptual tasks in minimally invasive scenes seeks to identify task combinations that can improve problem-solving accuracy \citep{simultaneous_depth_tool_seg,islam2020apmtl,scalable_joint_detection,multi_contour,multi_HOG,landmark_detection_submucosal} or efficiency, taking into account factors such as inference speed \citep{scalable_joint_detection,islam2020apmtl}, memory usage \citep{scalable_joint_detection,islam2020apmtl}, and annotation requirements \citep{scalable_joint_detection}. Furthermore, the existing body of literature demonstrates the presence of various distinct auxiliary tasks that can provide valuable guidance for optimizing perceptual tasks \citep{multi_contour,multi_HOG,landmark_detection_submucosal}.

In their work, \citet{simultaneous_depth_tool_seg} present an approach that concurrently tackles the depth estimation and binary tool segmentation tasks in laparoscopic images. The methodology employs a U-Net-like architecture with a shared encoder and two separate decoders, one dedicated to each task. Laparoscopic stereo images serve as the model's input.
In this framework, the segmentation decoder follows a conventional U-Net decoder style \citep{unet}. On the other hand, the depth estimation decoder generates disparity maps using the left-right consistency unsupervised depth estimation method, as outlined in \citet{monodepth17}. Notably, the unsupervised depth estimation approach is advantageous for endoscopic surgical datasets, which often lack ground-truth depth labels.
The results of this study indicate improvements in both instrument binary tool segmentation and unsupervised depth estimation tasks. This study suggests that binary tool segmentation and depth estimation could be effectively learned together within the MTL framework. However, additional datasets are required to generalize this hypothesis.

\citet{scalable_joint_detection} introduce an efficient and memory-friendly approach to MTL, which simultaneously addresses classification, instrument-type segmentation, and bounding box detection. The authors recognize the challenges associated with obtaining segmentation labels and, as a result, devise a solution that supports weak supervision.
This method harnesses the power of an EfficientNet backbone \citep{koonce2021efficientnet}. Unlike \citet{simultaneous_depth_tool_seg}, which advocates for distinct decoders for each task, the proposed model employs a single backbone with straightforward task heads for each objective. During training, the emphasis is placed on leveraging bounding box and classification labels, with only limited instrument segmentation labels. In cases where segmentation labels are unavailable, class activation maps are employed for guidance and supervision.
The outcomes reported by the authors are notable, demonstrating that even with just one per cent of the segmentation labels, they achieve commendable results in the surgical tool segmentation task which indicates that multitask learning could be used to reduce annotation costs. However, the results were evaluated on a single dataset EndoVis2018 \citep{endovis2018}, and more experiments are required to prove generalization.

\begin{figure}[t]
  \includegraphics[width=0.48\textwidth]{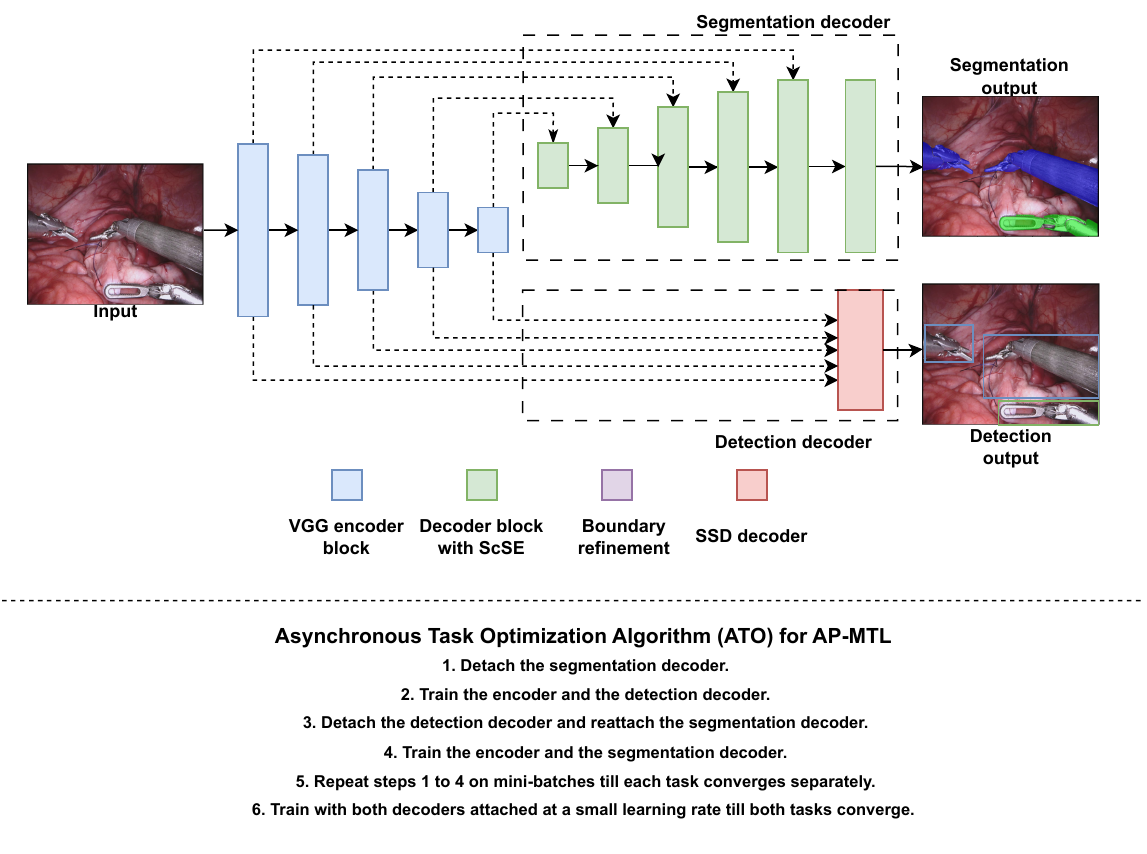}
  \caption{Illustration of the Attention Pruned Multitask Learning (AP-MTL) Network and the optimization method used for training this network \citep{islam2020apmtl}. The top image shows an encoder-decoder network with skip connections for its segmentation and detection decoders. A summary of the Asynchronous Task Optimization (ATO) for obtaining convergence for both tasks in the AP-MTL network is provided at the bottom.} 
  \label{fig:apmtl}
\end{figure}

In their work illustrated in \figref{fig:apmtl}, \citet{islam2020apmtl} introduce a real-time network designed for solving instrument detection and instrument-type segmentation in robotic surgery, referred to as the Attention pruned MTL network (AP-MTL). AP-MTL adopts a U-Net-like architecture with shared encoders and task-specific decoders. The decoder for object detection follows the design of the multi-box single shot detector (SSD) \citep{ssd_multibox}, while the segmentation network is a variant of the standard U-Net segmentation decoder. It incorporates custom squeeze excitation modules \citep{squeeze_excitation} known as the spatial channel squeeze excitation module (scSE). 
Additionally, \citet{islam2020apmtl} presents a custom optimization method known as Asynchronous Task-aware Optimization (ATO). It is a form of sequential training that optimizes different parts of the proposed network separately to ensure that correlated tasks converge at the same point, even if they do so at varying speeds. Following the optimization step, ATO introduces regularization to promote a more generalized network and ensures the smooth flow of gradients. To implement the ATO algorithm, gradients for each task are calculated and optimized separately, simulating an \emph{attach, optimize, and detach} mechanism for the task decoders. 
This study stands out by proposing a custom sequential training MTL optimization technique specifically tailored to their use case.

In their work, \citet{baby2023forks} introduce a modification to the standard Mask-RCNN and Mask2former framework \citep{he2017mask,cheng2022masked} for addressing instance segmentation in surgical instruments. The authors identify an issue where the classification of predicted masks often produces inaccurate results, despite the bounding box detection and mask segmentation tasks generally being performed correctly. To address this challenge, the authors propose the incorporation of a dedicated classification module to decouple the classification task from the region proposal and mask prediction processes. This new module takes as input multiscale features extracted from the feature extractor and the predicted instance masks. These predicted instance masks are employed for multiscale masking, and the results at different scales are subsequently merged and passed to a classification head. Sequential training is used where each stage of the proposed model is optimized separately with a training scheme. The authors report improvements compared to standard segmentation models and popular instrument segmentation models. However, with its three-stage approach, and extra parameters included compared to a standard Mask-RCNN and Mask2former, the inference speed and memory requirements would be increased.

\citet{zhao2022trasetr} propose TraSeTR, a method for leveraging tracking cues to enhance surgical instrument segmentation. This approach employs a transformer-based architecture that bears similarities to the DETR architecture \citep{carion2020end} for predicting instrument class, bounding box, and binary segmentation classes. \citet{zhao2022trasetr} improve on the standard DETR architecture by utilizing queries from prior frames for the instrument detection in the current frame. These queries are encoded with previous instrument information and serve as a form of tracking signal.  
These queries are applied to the transformer decoder, and identity matching between the previous queries and current queries are used as tracking cues. In addition, the authors also apply a contrastive query learning strategy to reshape the query feature space and alleviate difficulties in identity matching. The authors report improved performance on instrument segmentation using their approach. 
 This work builds upon ideas from DETR \cite{carion2020end} and Trackformer \cite{meinhardt2022trackformer}. TraSeTR can also be applied to non-surgical domains and compared to seminal works such as DETR and Trackformer, which would provide valuable insights.

Multitask learning for perceptual tasks has also been used for pituitary surgery.
\citet{das2023multi} present Pituitary Anatomy Identification Network (PAINet) and a newly introduced dataset for endoscopic pituitary surgery. They tackle the challenging task of identifying ten anatomical structures, with two prominent ones addressed through semantic segmentation and the other eight through centroid prediction. Their approach employs a U-Net architecture with two different task heads for each task for joint learning of the semantic segmentation and centroid prediction task. The authors report improvements in IoU and mean percentage of correct key points (MPCK) compared to standard non-multitask models.

These papers \citep{islam2020apmtl,scalable_joint_detection,simultaneous_depth_tool_seg,baby2023forks,zhao2022trasetr,das2023multi} demonstrate cases where the utilization of multiple tasks leads to improvements in the performance of all tasks.
They show that multitask learning can be highly efficient, achieving real-time performance while predicting multiple tasks. The use of weak supervision with multitask learning highlights the potential for efficient and cost-effective labelling strategies \cite{scalable_joint_detection}. However, it is important to note that these works do not report on negative transfer.
 
\citet{msdesis} highlights that MTL can face challenges as well. The authors present a method for jointly learning both disparity estimation and surgical instrument segmentation, but the results show a decrease in performance.
The architecture employed in \citet{msdesis} is U-Net-like, featuring a shared encoder and separate decoder heads for each task, including segmentation. The model undergoes pretraining using the Flythings3D dataset \citep{MIFDB16} for both the encoder and the disparity decoder.
Following pretraining, \citet{msdesis} present results from various training schemes. These include MTL training alone, training on one task and fine-tuning on the other, and training with MTL followed by fine-tuning on a single task.
Interestingly, the authors observed decreased performance when utilizing plain MTL for their two tasks. Instead, their experiments revealed that for marginal improvements in the segmentation task, it is necessary to first train the model using MTL and then fine-tune solely on segmentation. Conversely, for training the disparity task, it proves more effective to train the disparity task alone without segmentation. Due to the absence of a multitask dataset for depth and segmentation, multiple single-task datasets from different sources, were used for multitask training. Multitask learning not improving the disparity task, could also be because of the domain shifts.

\subsubsection{Perceptual tasks and auxiliary tasks}
In the previous papers, the focus is on enhancing the performance of two or more tasks by training them together.
However, an alternative approach is to prioritize one primary task and introduce another task as an auxiliary task \citep{multi_contour,multi_HOG,ss_depth_estimation_laparo,landmark_detection_submucosal}.
The role of the auxiliary task is to guide the training of the primary perceptual task.
These auxiliary tasks can take different forms. They can be real tasks with specific objectives \citep{ss_depth_estimation_laparo}, or derived tasks constructed from existing information in the image or labels \citep{landmark_detection_submucosal,multi_HOG,multi_contour}. In some cases, auxiliary tasks are used to inject domain-specific or prior knowledge into neural networks, enriching their capacity to learn and generalize \citep{landmark_detection_submucosal,ss_depth_estimation_laparo}.

In their work, \citet{multi_contour} introduce the concept of contour supervision, a form of boundary prediction to serve as an auxiliary task for semantic segmentation. Contour supervision involves the creation of object outlines or contours for class instances within a semantic segmentation, with the network tasked with predicting these contour maps.
The authors argue that contour prediction is valuable for localizing precise edges of the segmentation mask and providing information about the outer shape of objects. Different decoders are used to predict the contours and image segmentation. Contour supervision as an auxiliary task can be seen as a technique for improving the accuracy of image segmentation at object boundaries similar to loss functions that aim to improve models learning border pixels better \citet{unet}.

\begin{figure}[t]
  \includegraphics[width=0.48\textwidth]{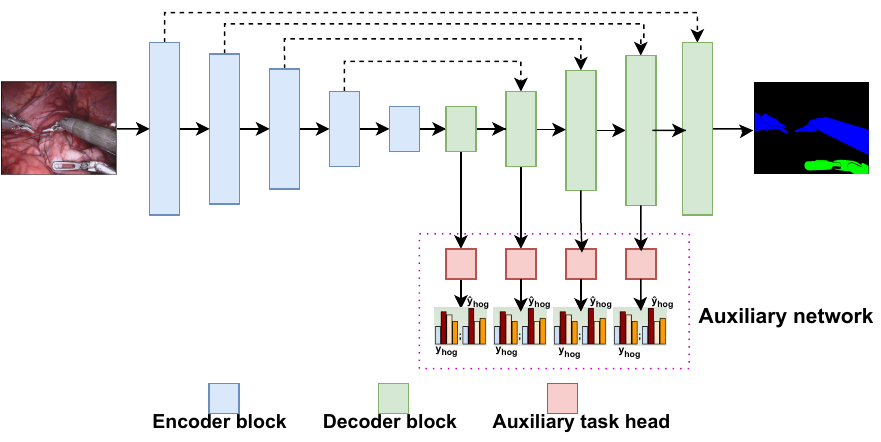}
  \caption{Histogram of gradient multitask learning (HoG-MTL) \citep{multi_HOG} demonstrates the use of an unconventional auxiliary task of predicting the histogram of gradients of input images image as an auxiliary task for semantic segmentation.} 
  \label{fig:hogmtl}
\end{figure}

In a different approach illustrated in \figref{fig:hogmtl}, \citet{multi_HOG} suggest employing the prediction of the Histogram of Gradients (HoG) of the image as an auxiliary task to guide the learning of the segmentation task. The authors contend that leveraging image features as pseudo-labels in image classification is a well-established practice, and the histogram of gradient is a widely used handcrafted feature for object detection. Therefore, learning HoG can serve as valuable supervision for the segmentation task.
To implement this approach, the authors propose a network that utilizes either the U2-Net \citep{Qin_2020_PR} or a U-Net with a single encoder and single decoder with deep supervision \citep{pmlr_v38_lee15a}. Auxiliary heads are integrated into the architecture to predict the histogram of gradients at various decoder levels.
This work provides further evidence that auxiliary tasks generated from the image can enhance the features extracted by a model.

\begin{table*}[htb!]
\caption{Methods that utilize multitask learning for tracking and control. *HP - hard parameter sharing. } \label{tracking_and_control:table}
\resizebox{\textwidth}{!}{
\begin{tabular}{|l|l|l|l|l|l|}
\hline
  & Publication & Tasks & Optimization & Architecture & Speed(fps)\\
\hline

\multirow{4}{*}{ \parbox{3.0cm}{Pose estimation}}  & \citet{concurrent_segmentation_iro_laina} & \thead[l]{tool segmentation\\2D pose heatmaps} & linear scalarization &  shared encoder with multiple  task branches (HP) & 18  \\
\bluecline{2-6}

 & \citet{art_net} & \thead[l]{tool presence detection\\ tool segmentation \\ geometric primitive detection} & linear scalarization &  shared encoder with multiple  task branches  (HP) & - \\
\bluecline{2-6}

 & \citet{multi_semi_pose} & \thead[l]{tool segmentation\\ tool number \\keypoint detection} & linear scalarization  &   shared encoder with multiple task branches (HP) & -\\
\bluecline{2-6}

 & \citet{li2021data} & \thead[l]{unsupervised depth estimation \\ tool-tip segmentation} & linear scalarization  & multistage network (HP) & 50 \\
\hline

\multirow{3}{*}{ \parbox{3.0cm}{Camera motion prediction}}  & \citet{learning_look_tracking} & \thead[l]{tool segmentation\\saliency prediction} & sequential training &  shared encoder with multiple  task branches (HP) & 127  \\
\bluecline{2-6}

 & \citet{Islam2020_ST_MTL} & \thead[l]{tool segmentation\\saliency prediction} & sequential training  &  shared encoder with multiple  task branches (HP) & 42 \\
\bluecline{2-6}

 & \citet{multi_fov} & \thead[l]{future optical flow \\ future tool segmentation \\ future camera action prediction} & linear scalarization  & multistage network (HP) & 23 \\

\hline

\end{tabular}
}
\end{table*}

In the paper \citet{ss_depth_estimation_laparo}, the authors focus on depth estimation and adopt the standard left-right consistency methodology, as introduced in \citet{monodepth17}. However, they introduce a novel auxiliary task by considering the 3D left-right consistency of 3D point clouds. The authors argue that while left-right consistency in disparity images is valuable, an additional source of information can be derived from the left-right consistency of point clouds, generated using predicted disparity images and information about laparoscopic stereo cameras.
Their method follows a two-stage MTL approach. In the first stage, a standard depth estimation architecture predicts 2D left and right disparity images. Subsequently, these disparity images, along with the focal length, the stereo distance, and predetermined blind masks for outlier removal, are used to generate 3D point clouds. The 2D depth loss functions include an appearance matching loss, a smoothness loss, and a left-right disparity consistency loss. In addition, the authors employ the iterative closest point (ICP) algorithm \citep{besl1992method} to create a loss function for 3D left-right consistency by using the final residual registration error after ICP minimization. These losses are combined using linear scalarization.
Incorporating the 3D auxiliary task is a key feature, as it enables the integration of information about the stereo camera into the optimization process of the neural network.

In endoscopic submucosal dissection (ESD), dissection landmarks are crucial for marking the boundary between a lesion and normal tissues. ESD involves creating landmarks around the lesion to label the boundary between the lesion and normal tissues \citep{ono2021guidelines}. \citet{landmark_detection_submucosal} present a neural network designed for detecting dissection landmarks.
This approach stands out by introducing an auxiliary task for capturing the spatial relationship between each dissection landmark. Essentially, the authors seek to enhance the detection of dissection landmarks by incorporating domain knowledge that dictates alignment along a curve. To represent the spatial relationships between the nearest landmark neighbours on the curve, an edge map is generated. This edge map is then transformed into a heatmap that preserves these spatial relationships.
The authors propose a shape-aware relation network based on the U-Net architecture featuring multiple decoders. This network functions as an MTL system that predicts both landmark positions and the heatmap of landmark spatial relationships. Similar to \citet{ss_depth_estimation_laparo}, auxiliary tasks were used to incorporate domain knowledge into the network, yielding positive results.

\subsection{MTL for tracking and control} \label{sec::mtl_surgical::pose_automatic_camera}

In this subsection, we explore multitask learning (MTL) applications in the context of tracking in the surgical scene and the control of instruments or minimally invasive cameras. We focus on two main areas within this context: pose estimation and camera motion prediction. Pose estimation involves determining the position and orientation of surgical instruments or anatomical structures within the surgical field. Camera motion prediction, on the other hand, deals with forecasting the correct movement of the minimally invasive camera based on the current and past states of the surgical environment.

\tabref{tracking_and_control:table} summarizes the papers discussed in this subsection, highlighting the multiple tasks addressed, the optimization approaches adopted, the architectural strategies used for multitask learning, and the reported speeds for each study.

\subsubsection{Pose estimation}
Pose estimation is the process of determining the spatial configuration of objects or entities within a given scene, with a focus on estimating their positions and orientations. Pose estimation can be categorized into two branches based on the coordinate system in which the pose is measured: 2D pose estimation and 3D pose estimation.
2D pose estimation is particularly well-suited for scenarios where depth information is not essential or available. The goal is to accurately localize key points or landmarks that define the orientation of the object of interest within the image.
On the other hand, 3D pose estimation involves estimating the full three-dimensional spatial configuration of objects in the scene. This estimation includes both the 2D position and depth information, along with orientation, making it a more challenging task.
Accurate pose estimation is of paramount importance in various applications related to endoscopic surgeries, including instrument tracking, automatic surgical camera control, augmented reality applications, and robotics. It is worth noting that the literature in the field of endoscopic surgeries often uses the 2D coordinate space \citep{multi_semi_pose,concurrent_segmentation_iro_laina,art_net,learning_look_tracking,Islam2020_ST_MTL,li2021data}. 
Still, it may use algebraic and projective geometry techniques or depth information to derive 3D pose estimation if needed \citep{art_net,li2021data}.

In their paper, \citet{concurrent_segmentation_iro_laina} introduce a 2D instrument pose estimation method, employing the MTL paradigm. The two tasks trained for this purpose are segmentation and a variant of the instrument keypoint detection task. To achieve this, the authors reframe the 2D pose landmark detection task as a regression problem focusing on instrument heatmaps.
The model utilized in this approach is a variant of the U-Net-like architecture with two different decoders: a segmentation decoder and a heatmap regression decoder.  The output from the segmentation task is combined with the feature maps in the heatmap regression branch, a design choice that contributes to the improvement of predicted regression maps. This work suggests that keypoint heatmap prediction and tool segmentation might be complementary tasks. However, more datasets and experiments are needed to generalize this finding.

\begin{figure}[t]
  \includegraphics[width=0.48\textwidth]{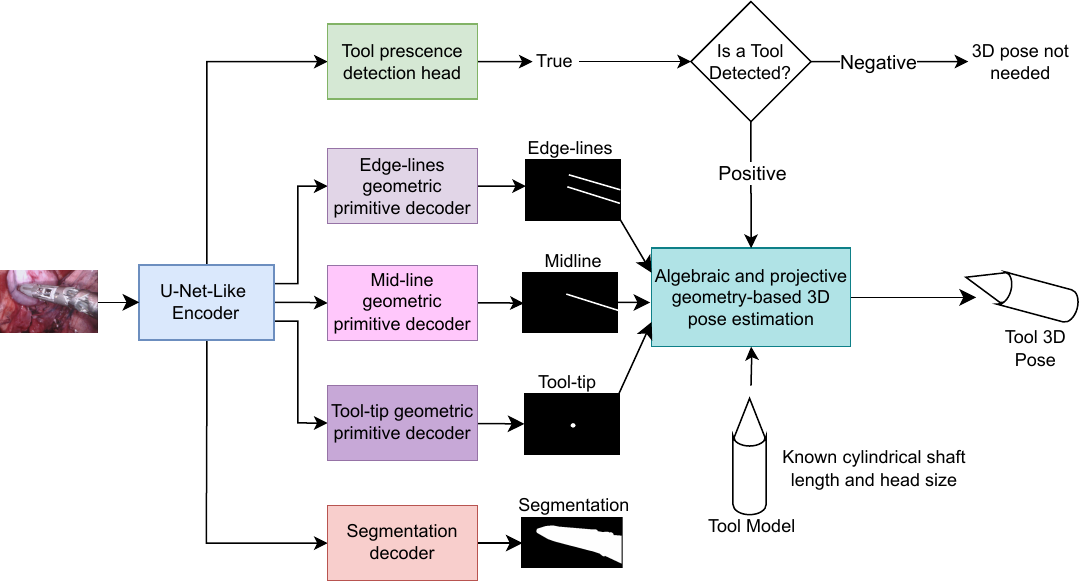}
  \caption{The Augmented Reality Tool Network (ART-NET) \citep{art_net} presents a system to produce 3D pose estimations of instruments for augmented reality and 3D length measurement applications. As seen in the diagram, ART-Net simultaneously learns to predict tool presence, edge-lines, mid-lines, tool-tip and segmentation. The geometric primitives predicted are combined with prior information to produce the required Tool 3D pose.} 
  \label{fig:artnet}
\end{figure}

\citet{art_net} present an MTL method for instrument presence detection, segmentation, and 2D pose estimation. The 2D pose estimations are represented as `geometric primitives', which in their work refer to heat maps of critical components of the instruments, such as edge lines, mid-lines, and the tool-tip, as depicted in  \figref{fig:artnet}. The geometric primitives approach is chosen as they easily facilitate the calculation of 3D instrument poses.
Unlike conventional 2D pose estimation methods that predict landmark positions \citep{multi_semi_pose} or heatmaps ~\citep{concurrent_segmentation_iro_laina}, this approach predicts a geometric primitive map of the relevant tool parts, including the tool edge, shaft mid-line, and tool-tip. The network architecture employed is U-Net-like, featuring a single encoder, direct instrument presence detection connected to the encoder, and four decoders for segmentation, edge-line primitive mapping, shaft mid-line primitive mapping, and tool-tip primitive mapping.
The 2D geometric primitive map, once predicted, is then used in combination with prior information about the instrument, such as the radius of the tool shaft and the length of the tool head, to calculate the required 3D pose using algebraic and projective geometry techniques, followed by refinement. 
With multiple decoders for each geometric primitive, the network was very slow and unsuitable for real-time applications. To address this, the authors used depth-wise separable convolutions \citep{chollet2017xception}. Exploring ways to reduce the number of decoders while preserving geometric primitives could further enhance the speed of this model.

\citet{multi_semi_pose} introduce an alternative method for predicting 2D pose estimation while leveraging the MTL paradigm. The authors highlight a common challenge in the field - the abundance of datasets for tasks like surgical instrument segmentation, contrasted with the scarcity of datasets for pose estimation. To address this issue, they propose a multitask semi-supervised approach to pose estimation. Their method predicts three different outputs: instrument segmentation, instrument number and instrument landmark keypoint detection. The instrument segmentation task is used as a spatial constraint, and the instrument number prediction task as a global constraint for the instrument landmark keypoint detection task. The semi-supervised technique employed here is the mean-teacher framework for semi-supervised learning \citep{tarvainen2017mean}. Both student and teacher models take the form of feature pyramidal networks \citep{lin2017feature}, each with task-specific heads for the three tasks trained in conjunction.  

\citet{li2021data} introduce a three-stage framework for optimizing laparoscopic field of view (FOV) control during minimally invasive surgery (MIS) through tracking and 3D localization. The framework begins with 3D localization, using multitask learning with a shared encoder and task-specific decoders for simultaneous surgical tool-tip segmentation and depth estimation. The predicted tool-tip segmentation and depth, along with physical constraints like tool diameter, generate the 3D pose of the tool-tip.
In the optimal view control stage, the framework tracks the localized tool-tip and focuses on a data-driven 2D image region to perform laparoscopic control. The final stage includes an affine mapping-based Minimize Rotation Constraint (MRC) method to correct visual misorientation from the Remote Center of Motion (RCM) constraint, and a null-space controller to optimize 2D and 3D tool positions relative to the laparoscope. 
This work uses a heuristic of tracking the tool-tip for camera control. However, the camera motion becomes reactive, and this heuristic can fail in complex scenes with multiple tools moving in and out of the operating field.

As evident from the literature on instrument/camera tracking methods in MIS, such as \citep{art_net,multi_semi_pose,concurrent_segmentation_iro_laina,li2021data}, the predominant approach involves predicting some form of 2D pose estimation rather than 3D pose estimation. This predicted 2D pose is usually used to facilitate other tasks such as tracking by prediction of 2D poses in a video sequence, as exemplified in \citet{concurrent_segmentation_iro_laina}, generating 3D poses to be used for laparoscope automation via tracking \citep{li2021data} or enabling augmented reality overlays, as demonstrated in \citet{art_net}.

Another closely related and prevalent task to surgical instrument pose estimation is motion prediction for cameras, which also plays a significant role in the context of tracking and control in minimally invasive surgeries.

\subsubsection{Camera motion prediction}
The motion prediction task for cameras has been explored along two primary avenues: scanpath prediction and camera imitation learning.
Scanpath prediction involves forecasting the potential camera paths based on the most captivating elements within the camera's current field of view. It draws inspiration from a theory in human gaze fixation, which posits a direct connection between human gaze scanpath and the attentional priority of objects in an image, often referred to as object saliency. In simpler terms, humans tend to focus on the most vital information first before shifting their gaze to less significant details. In the context of MIS, scanpath prediction tasks are centred around predicting the most salient objects in a scene and plotting a path from the most salient to the least salient areas.
\citet{Foulsham} provide a good reference for further insights on scanpaths, saliency, and their correlations.

\begin{figure}
  \includegraphics[width=0.48\textwidth]
   {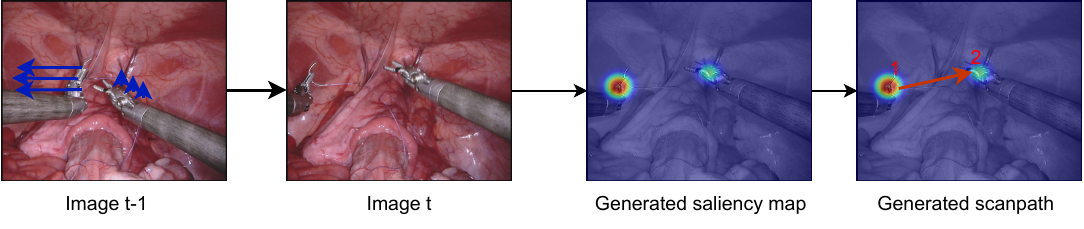}
  \caption{Illustration of the heuristic used for the generation of saliency maps and scanpaths in \citet{learning_look_tracking}.
  A saliency map is generated (using the code provided by \citet{learning_look_tracking} and images from EndoVis2017 \citep{endovis2017}), which correlates to the motion of fixation points and the size of the fixation points, which are assumed to be located on the instrument wrist and claspers. The scanpath is then assumed to be the movement between the most salient to the least salient instrument. }
  \label{fig:scanpath_mis}
\end{figure}

\citet{learning_look_tracking} introduce a method for estimating saliency in surgical scenes, specifically by training in conjunction with segmentation as an auxiliary task. A notable challenge in this context is the absence of ground truth saliency datasets for MIS. To address this issue, the authors propose a novel method for generating saliency maps.
The approach assumes that the most intriguing parts of surgical instruments are the wrist and claspers, and fixation points are located on these instrument segments. Additionally, the movement of instruments is utilized as a key indicator of saliency. Instruments with more significant movement are assigned higher saliency values than those with minimal movement. A scanpath is then defined as the movement from the most salient to the least salient instrument.
\figref{fig:scanpath_mis} illustrates the process of generating a scanpath from images as presented in \citet{learning_look_tracking}.
The architectural framework employed is based on a U-Net structure featuring an encoder and two decoders, one dedicated to saliency map prediction and the other to instrument class segmentation. Attention modules are incorporated in the saliency map prediction branch to suppress irrelevant regions and highlight salient features. The authors utilize a two-phase learning strategy to address the challenge of converging both tasks in the same epoch. In the first phase, equal loss weights are assigned to both tasks. In the second phase, the model fine-tunes the loss weights based on the performance of the converged task. \citet{learning_look_tracking} is unique in MTL for MIS it diverges from the standard linear scalarization optimization strategy.

\begin{table*}[htb!]
\caption{Methods that utilize multitask learning for surgical video workflow analysis. *HP - hard parameter sharing. } \label{workflow:table}
\resizebox{\textwidth}{!}{
\begin{tabular}{|l|l|l|l|l|l|}
\hline
  & Publication & Tasks & Optimization & Architecture & Speed(fps)\\
\hline

\multirow{6}{*}{ \parbox{3.0cm}{Phase and instrument detection}}  & \citet{EndoNet} & \thead[l]{phase recognition\\tool presence detection} & linear scalarization & multistage network (HP) & - \\
\bluecline{2-6}

 & \citet{twinanda2016lstm} & \thead[l]{phase recognition\\tool presence detection} & linear scalarization & multistage network (HP) & - \\
\bluecline{2-6}

 & \citet{tecno} & \thead[l]{phase recognition\\tool presence detection} & linear scalarization  & multistage network (HP) & - \\
\bluecline{2-6}

 & \citet{multitask_connectionism} & \thead[l]{phase recognition\\tool presence detection} & linear scalarization  & multistage network (HP) & -  \\
\bluecline{2-6}

& \citet{Sanchez-Matilla2022_data_centric} & \thead[l]{phase recognition\\scene segmentation} & sequential training  & multistage network (HP) & - \\
\bluecline{2-6}

& \citet{multi_task_recurrent_conv} & \thead[l]{phase recognition\\tool presence detection} & sequential training  & multistage network (HP) & 3.3\\
\bluecline{2-6}

\hline

\multirow{6}{*}{ \parbox{3.0cm}{Surgical action triplet learning}}  & \citet{1nywoye_rec_action_triplets} & surgical action triplet recognition & linear scalarization &  shared encoder with multiple  task branches (HP) & -  \\
\bluecline{2-6}

& \citet{NWOYE2022102433_rendevous} & surgical action triplet recognition & uncertainty weighting &  shared encoder with interacting task branches  (HP) & 28.1 \\
\bluecline{2-6}

& \citet{sharma2022RendezvousInTime} & surgical action triplet recognition & linear scalarization &  shared encoder with interacting task branches (HP) & - \\
\bluecline{2-6}

& \citet{multi_self_distillation} &  \thead[l]{surgical action triplet recognition\\phase recognition}  & linear scalarization &  encoder with multiple branches (HP)  
 & - \\
\bluecline{2-6}

& \citet{sharma2023surgical} & surgical action triplet detection & linear scalarization & multistage network (HP) & - \\
\bluecline{2-6}

\hline

\multirow{2}{*}{ \parbox{3.0cm}{Multi-granularity activity detection}}  & \citet{Ramesh2021_phase_step_gastric_bypass} &  \thead[l]{phase recognition \\ step recognition} & linear scalarization & multistage network (HP) & - \\
\bluecline{2-6}

& \citet{holistic_scene_understanding} & \thead[l]{phase recognition \\ step recognition \\ Action recognition \\ Instrument detection} & linear scalarization &  multiple encoders with  with multiple task branches (HP) & - \\
\bluecline{2-6}
\hline

\end{tabular}
}
\end{table*}

Building upon the foundation laid by \citet{learning_look_tracking}, the subsequent paper on scanpath prediction with MTL, known as ST-MTL \citep{Islam2020_ST_MTL}, maintains a similar architectural framework. However, the authors assert that the correct prediction of saliency maps requires information from the current frame and insights from previous frames. To address this temporal relationship, they introduce a convolutional Long Short-Term Memory (ConvLSTM) \citep{ConvLSTM} into the saliency prediction decoder.
The utilization of ConvLSTM modules enhances the model's ability to capture temporal dependencies and leverage information from past frames, contributing to more accurate saliency predictions. Moreover, \citet{Islam2020_ST_MTL} incorporate a sequential training method for a spatiotemporal model similar to ATO optimization method developed in \citet{islam2020apmtl}, as discussed in \secref{section::perceptual}.

Scanpath prediction, despite its merits, has several limitations in minimally invasive surgery (MIS). It currently focuses solely on surgical instruments, ignoring critical tissues and organs, which can lead to suboptimal camera movements. The dynamic nature of surgical procedures, with instruments frequently entering and exiting the field of view, poses challenges for scanpath models to adapt quickly. Additionally, the complexity of the surgical environment, involving multiple instruments and varying tissue types, can be oversimplified by saliency prediction. These models also lack contextual awareness, failing to consider the surgeon's intentions or the stage of the operation, which can result in misalignment with the surgeon's preferred view and potential inefficiencies.

Another promising avenue in the realm of camera motion prediction is camera imitation learning. This approach emphasises emulating the camera movements observed during surgeries performed by surgeons or their assistants. This technique is particularly relevant in the context of MIS, which often follows predefined procedural steps. These surgeries typically adhere to a fixed sequence of steps, and under normal circumstances, camera movements should exhibit similar patterns. Camera imitation learning aims to address the reactive nature of instrument tracking for camera control, as seen in \citet{li2021data,gruijthuijsen2022robotic}, by adding contextual awareness and emulating the behaviour of surgical assistants in various scenarios.

\citet{multi_fov} introduce a method for the proactive adjustment of the camera's field of view, achieved by modifying the camera's position in the $x$, $y$, and $z$ coordinates.  To replicate the motion patterns of a laparoscope camera, the authors employ a ConvLSTM for sequence modelling, predicting feature motions on a per-frame basis.
A unique aspect of this research is the generation of ground truth laparoscope motion from laparoscopic videos. This process involves dynamic camera motion estimation to infer camera pose.
The authors utilize the Neural-Guided Random Sample Consensus (NG-RANSAC) method~  \citep{brachmann2019neural} to match stereo-images under dynamic conditions.
Additionally, they apply the remote centre of motion constraint to optimize the pose estimation. The ground truth motion in the $x$, $y$, and $z$ directions is derived from different sequential poses.
The inputs to the ConvLSTM model comprise estimated segmentation and optical flow obtained from off-the-shelf models. The model optimizes for correct outputs of the subsequent $N$ optical flow and segmentation results. The optical flow outputs are then passed through a laparoscopic action head to predict the camera's movements in the $x$, $y$, and $z$ coordinates. \citet{multi_fov} improve on their previous work \citep{li2021data} by focusing on proactive camera motion and learning the motion strategy from clinical videos instead of reactive motion prediction and instrument tracking. Additionally, \citet{multi_fov} directly predict in 3D space instead of generating results in 2D and scaling to 3D, making their approach unique among the methods reviewed.

\subsection{MTL for surgical video workflow analysis}
Surgical video workflow analysis is the systematic examination of videos recorded during surgical procedures, aiming to extract valuable insights into various facets of the surgery. This analysis serves multiple critical purposes, including providing context-aware intraoperative assistance, enhancing surgeon training, facilitating procedure planning, supporting research endeavours, and enabling retrospective analysis \citep{lalys2014surgical}.

The analysis process involves breaking down a surgical procedure video into distinct segments, which are categorized based on the surgeon's various activities. Different surgeries can be divided into specific activities, which can be examined at multiple levels of granularity. 

\begin{figure}[t]
  \includegraphics[width=0.48\textwidth]{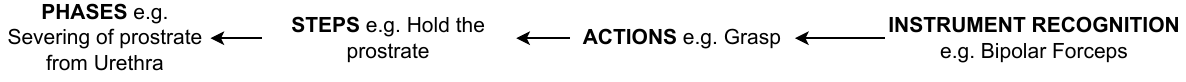}
  \caption{A visual representation of the different levels of granularities in surgical video workflow analysis. This representation for surgical video workflow analysis is adopted from the framework for holistic analysis of surgical videos in \citet{holistic_scene_understanding}. } 
  \label{fig:phase_relationships}
\end{figure}

We adopt a representation for surgical video workflow analysis from \citet{holistic_scene_understanding}, where the holistic analysis of a surgical video and its workflow is divided into four granularities. 
The initial granularity level involves identifying \emph{instruments} present in the surgical scene, which are features of the surgical scene directly responsible for the surgical workflow. 
This stage entails understanding the instruments within the field of view as these instruments movements and interactions with different tissues give rise to specific \emph{actions}, such as grasping or cutting, which represents the second granularity level.
The third granularity stage involves the sequence of actions performed in pursuit of a particular surgical objective, which is called a \emph{step}. Multiple steps are executed together to accomplish a portion of the surgery, termed a surgical \emph{phase}, which is the fourth and final granularity of a surgical procedure.
\figref{fig:phase_relationships} provides a visual representation of the different levels of granularity. 

Addressing the problem of surgical video workflow analysis is a critical step towards enhancing computer-aided interventions in surgical procedures. However, recognizing and distinguishing between different phases, steps, or actions within surgical videos remains a formidable challenge. This challenge is attributed to several factors, including the limited availability of publicly accessible data, the high resemblance between long-range sequences belonging to different phases or steps, the substantial variability within sequences associated with a single phase or step, and the extended duration of surgical procedures necessitating thorough analysis.

In this section, we explore the literature which applies MTL to solve the surgical activity recognition problem and divide it into three subsections, namely: auxiliary tasks for surgical activity recognition, surgical action triplet learning, and multi-level activity recognition. 
\tabref{workflow:table} summarizes the papers discussed in this subsection, highlighting the multiple tasks addressed, the optimization approaches adopted, the architectural strategies used for multitask learning, and the reported speeds for each study.

\subsubsection{Phase and instrument detection}
Early research in surgical activity recognition with MTL primarily centred on phase recognition \citep{EndoNet,tecno,Sanchez-Matilla2022_data_centric,multi_task_recurrent_conv}. 
phase recognition was usually predicted alongside instrument recognition specifically tool presence detection. The rationale for focusing on instrument recognition is rooted in the understanding that surgical instruments are the means by which surgeons interact with the surgical scene. Assuming standard surgical practices are followed, a substantial correlation exists between the choice of specific tools and the corresponding activities undertaken during a surgical procedure.

\begin{figure}
  \includegraphics[width=0.48\textwidth]{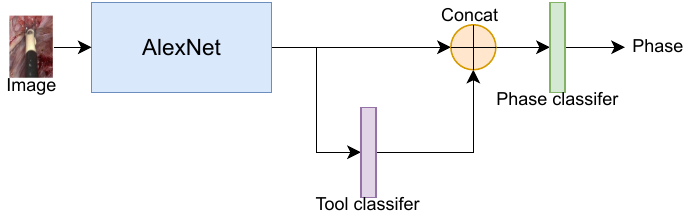}
  \caption{One of the earliest works that attempts to solve the phase recognition problem is the EndoNet \citep{EndoNet}. It uses an Alexnet as a feature extractor with two task heads - A tool classifier head and a phase classifier head. Logits from the tool classifier head are concatenated with features from Alexnet before predicting phase.} 
  \label{fig:endonet}
\end{figure}

One of the pioneering papers that introduces the application of MTL and CNNs to surgical phase recognition is EndoNet~\citep{EndoNet}. This method follows a two-stage prediction approach. The first stage jointly learns tool presence and phase prediction for each frame. The second stage refines the phase predictions generated in the initial stage. To implement this, EndoNet employs a pretrained AlexNet \citep{alexnet} as the encoder for its first stage, which is fine-tuned by simultaneously predicting tool presence and phase detection using dedicated task heads.
In the second stage, EndoNet combines the features for phase detection in the current frame with information from the previous frames and feeds this combined data into a Hidden Markov Model (HMM) \citep{HHMM}. An architectural overview of EndoNet is illustrated in \figref{fig:endonet}. In another approach, \citet{twinanda2016lstm} replaces the HMM in EndoNet with an LSTM \citep{LSTM} to improve temporal modelling. Incorporating a temporal model, such as HMM in EndoNet and LSTM in \citet{twinanda2016lstm}, significantly improved phase recognition, validating the intuition that temporal features are crucial for accurate phase prediction.
\citep{EndoNet} released the Cholec80 dataset for benchmarking phase recognition and tool presence detection using accuracy, precision, and recall as metrics.

In a manner similar to the approach employed by \citet{EndoNet}, \citet{tecno} also leverage the tool presence detection task along with a multistage network for phase recognition with different temporal modelling and optimization. Instead of utilizing the LSTM/HMM models as seen in previous works, they opt for the Temporal Convolutional Network (TCN), as described by \citet{TCN}, for the refinement phase.
The multitask optimization method used involves implementing the median frequency balancing technique \citep{pred_depth_normals_sem}. As for the input to the TCN, it comprises a concatenation of features extracted from the current frame and the $N$ preceding frames. TCN is preferred to LSTM/HMM approaches because it is notably faster, less resource-intensive and gives competitive results.

In a similar vein, \citet{multitask_connectionism} delve into the intricate relationship between tool presence detection and phase detection. The authors introduce a two-stage network identical to previous approaches, but \citet{multitask_connectionism} differ in how they link the phase detection task with the tool presence detection task.
Unlike the methodologies found in \citep{tecno,twinanda2016lstm,EndoNet}, where refinement primarily concerns phase detection, this approach extends refinement to both tool and phase detection.
Key to their methodology is the introduction of a joint probability loss function, which serves as the binding agent between the two tasks at each stage. The joint probability loss function is a product of the probabilities associated with tool and phase detection, weighted by the inverse of the tool's appearance frequency in a given phase. This approach seeks to establish a more integrated and interdependent relationship between the tasks, offering a fresh perspective on the problem. A training procedure was designed that sequentially trains different parts of the multistage network with different losses and then jointly optimises the whole network.

\citet{Sanchez-Matilla2022_data_centric} take a distinct approach by strongly emphasising enhanced temporal modelling by adopting a different task from tool presence detection for their two-stage network. In this paper, the authors opt for semantic segmentation and phase recognition, deviating from the conventional choice of tool presence detection.
The first stage is a multitask network incorporating two distinct decoders, one for segmentation and another for phase prediction. The second stage of their network implementation employs a Temporal Convolutional Network (TCN) for the refinement phase. A notable finding of their research is the potential for performance enhancement with as little as 5\% of the data labelled for segmentation.

In their research, \citet{multi_task_recurrent_conv} offer an end-to-end trained network that capitalizes on the strong correlation between tool presence and surgical phases, much like previous works. However, it distinguishes itself by adopting a single-stage network design.
The authors introduce a novel network known as MTRCNet-CL, featuring a shared encoder with two branches. One branch is dedicated to tool presence detection, while the other is focused on phase recognition. The phase recognition decoder branch incorporates an LSTM for temporal modelling.
An innovative aspect of their approach is formulating a correlation loss that models the intricate relationship between tool presence and the predicted phase in the decoders. Their hypothesis posits that given the correlation between these two tasks, the logit values of the two tasks should also exhibit a certain degree of correlation. To measure this correlation, they calculate the Kullback-Leibler (KL) divergence between the phase logit values and the tool presence logit values.

\citet{EndoNet,multi_task_recurrent_conv,multitask_connectionism,tecno,Sanchez-Matilla2022_data_centric} were all benchmarked using the Cholec80 dataset and similar metrics, facilitating direct comparisons between these works. This highlights the benefits of using a multitask dataset with predefined tasks and widely accepted metrics, as shown in Table \ref{cholec80_auxilliary:table}.

\begin{table}[htb!]
\caption{Phase recognition and instrument detection results of various multitask learning methods on Cholec80. Results are retrieved from references. The mean of the published results are reported.} \label{cholec80_auxilliary:table}
\resizebox{\columnwidth}{!}{
\begin{tabular}{|l|l|l|l|l|}
\hline
 & \multicolumn{3}{|c|}{\textbf{Phase Recongition}} & \textbf{Tool presence detection}  \\
\hline
 
 Publication &  Avg. precision & Avg. Recall & Accuracy & Avg. precision  \\
\hline
EndoNet\citet{EndoNet} & 81.7 & 73.7 & 79.6 & 81.0 \\
\citet{Sanchez-Matilla2022_data_centric} & -  & -  & 89.51  & -  \\
TeCNO\citet{tecno} & 88.56  & 81.64  & 85.2  & -  \\
\citet{multitask_connectionism} & 85.7 & 83.5 &  92.0  & 93.53  \\
MTRCNet-CL \citet{multi_task_recurrent_conv} & 86.9 & 88.0 &  89.2  & 89.1 \\
\hline
\end{tabular}
}
\end{table}

As evident from the studies presented in this section, early research in surgical activity detection primarily concentrated on the surgical phase granularity. Researchers explored various architectural and loss function strategies to leverage the relationship between phase recognition and tool presence detection. However, it is important to note that tool presence detection has inherent limitations. It can introduce noise into the analysis and lacks the precision required to pinpoint where the surgical action is unfolding within the video.
Moreover, beyond the tools, the specific tissues that surgical tools interact with during a procedure also offer valuable insights into the ongoing surgical activity. 
Additionally, the identification of the instrument, tissue, and the actions taking place in a scene can provide valuable clues about the ongoing surgical steps and phases \citep{holistic_scene_understanding}. 
Each step and phase requires very specific actions performed on specific tissues, in a specific order, and with specific instruments.

\subsubsection{Surgical action triplet learning}
\begin{figure}[htb!]
   \centering
  \includegraphics[width=0.48\textwidth]{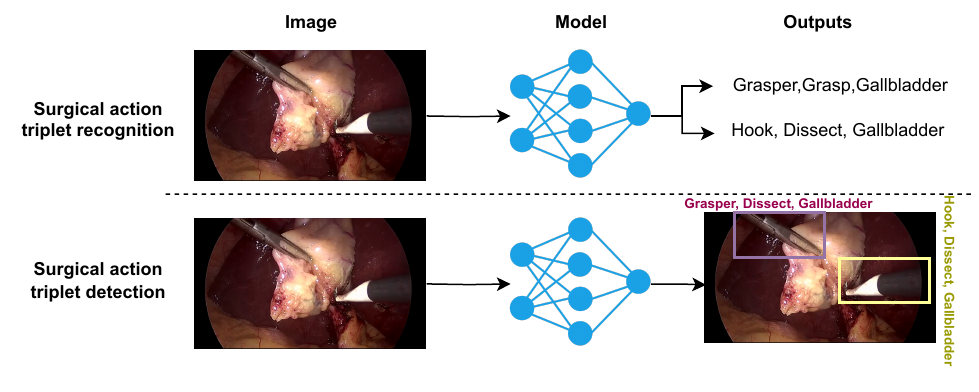}
  \caption{The triplet task of recognizing the $<$instrument, action and target$>$ is traditionally cast as surgical action triplet recognition, which is a multi-label multi-classification recognition problem \citep{1nywoye_rec_action_triplets,NWOYE2022102433_rendevous,sharma2022RendezvousInTime} as shown in the top image. Recently, there has been a progression in the task difficulty that cast this problem as surgical action triplet detection, which involves localizing all instruments and associating the corresponding instrument triplet to all localized instruments \citep{sharma2023surgical}.} 
  \label{fig:triplet_example}
\end{figure}

More recent literature has compellingly argued against the exclusive prediction of surgical workflow either directly or solely based on tool presence. It has been contended that such an approach proves inadequate for a comprehensive understanding of surgical scenes. Instead, these works shift their focus towards the more granular action recognition task, striving to establish connections between three crucial elements before attempting to predict activities with longer sequences: the specific instrument in use, the action being performed, and the target anatomy undergoing the procedure, often collectively represented as three distinct labels, $<$instrument, action, target$>$.
This predictive task illustrated in \figref{fig:triplet_example} and involving these three distinct labels is often referred to as the `surgical triplet prediction task' \citep{laproscopic_rule_based_action_triplets}.
The surgical action triplet task fundamentally represents a multilabel (instrument, action, target) multiclass classification problem, with the understanding of the relationship between the labels being paramount. Typically, the problem is framed with the input as a single frame \citep{1nywoye_rec_action_triplets,NWOYE2022102433_rendevous}, or multiple consecutive frames \citep{sharma2022RendezvousInTime}, with the objective of predicting the $<$instrument, action, target$>$ triplet. 
The benchmark datasets for surgical triplet prediction are CholecT45 and its more recent version, CholecT50, which includes five additional sequences \citep{NWOYE2022102433_rendevous}. The surgical action triplet tasks are evaluated using IVT metrics:

\begin{enumerate}
    \item Component Average Precision: This evaluates the correct recognition of the instrument ($AP_I$), verb ($AP_V$), and target ($AP_T$) components of the triplets.
    \item Triplet average precision: This assesses the correct recognition of interactions between tools, actions, and targets, including instrument-verb interaction ($AP_{IV}$), instrument-target interaction ($AP_{IT}$), and the main metric, instrument-verb-target ($AP_{IVT}$), for surgical action triplet recognition
\end{enumerate}

For more information about the metrics behind the CholecT45 and CholecT50 datasets, we refer readers to \citet{nwoye2022data}. All the works in this subsection use the CholecT45, CholecT50, or extensions of these datasets with additional information (such as instrument bounding boxes). The results of the studies on surgical action triplet prediction are presented in \tabref{triplet_results:table}.

In their study, \citet{1nywoye_rec_action_triplets} train a multitask network tailored specifically for the surgical action triplet task. They proposed the Tripnet architecture, which employs an encoder to extract joint features and three dedicated decoders for each of the task components.
The first decoder operates as a convolutional-based unit, with a dual purpose: it predicts instrument classes in the image and generates class activation maps. These class activation maps are essential for the functioning of the other two decoders. The second and third decoders are \emph{class-activation-map-guided} convolutional units that extract features relevant to action recognition and target recognition, respectively. They use the class activation maps from the instrument decoder as a guide.

\begin{figure}[t]
   \centering
  \includegraphics[width=0.48\textwidth]{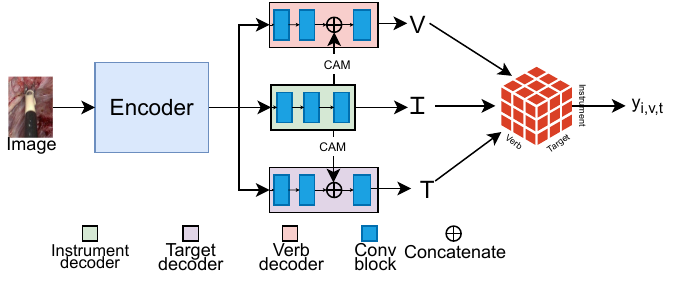}
  \caption{The proposed Tripnet architecture for surgical action triplet recognition \citep{1nywoye_rec_action_triplets}. The model contains an encoder and three separate branches for predicting instrument verb and targets. The class activation maps from the instrument branch are used as weak localization guides for action and target prediction.} 
  \label{fig:tripnet}
\end{figure}

The class activation maps are concatenated with the features in the action and tissue target decoders. This approach is underpinned by the hypothesis that the instrument plays a pivotal role in interacting with the surgical environment to initiate actions on a desired target. By incorporating a weak localization guide indicating the instrument's current location, the authors expect to enhance the performance of both the action recognition and tissue target recognition tasks. A visual representation of the Tripnet structure can be found in \figref{fig:tripnet}.

Furthermore, the authors observe that not all triplet combinations are feasible, and a data association problem arises as these components (predicted instruments, actions, and tissues) are interconnected. Instead of predicting each label separately, the authors opt for a joint prediction approach. The logits produced from each encoder ($I$ for Instrument, $V$ for action verb, and $T$ for tissue target) are used to create a 3D interaction space volume ($Y$) through an outer product operation:
$$
    Y = \alpha I \otimes \beta V \otimes \gamma T 
$$
Here, $\alpha$, $\beta$, and $\gamma$ represent learnable weights. The 3D volume is quantized, with values above a chosen threshold accepted as valid triplets, while spaces in the 3D volume that can never form triplets are masked out. The loss function for this approach comprises the standard cross-entropy loss for all tasks, combined into a linear combination of losses.

\citet{NWOYE2022102433_rendevous}, also known as RDV, is a sequel to the earlier work \citep{1nywoye_rec_action_triplets}.
This paper introduces a noteworthy improvement by incorporating repeated attention mechanisms to facilitate inter-task knowledge transfer, drawing inspiration from the transformer architecture. Similar to \citet{1nywoye_rec_action_triplets}, RDV employs a single encoder with three decoders. However, it distinguishes itself by employing attention-based decoders, referred to as the \emph{Class Activation Guided Attention Mechanism} (CAGAM).
 
The CAGAM mechanism replaces class-activation-map-guided convolutional units that extract features relevant to action recognition and target recognition used in the Tripnet architecture \citep{1nywoye_rec_action_triplets}. Similar to the Tripnet architecture, It leverages class activation maps (CAM) from the Instrument decoder as a source of weakly supervised spatial guidance for the action and tissue target recognition tasks. But instead of just concatenating, the CAM is used to produce Key and Query vectors to form an attention matrix, which is used to scale the features from the Targets and Actions. 
Following the CAGAM Mechanism, RDV yields instrument, action, and target features, which are further processed through another attention mechanism. This mechanism is a variant of the multi-head attention models used in transformers. This process is integral to combining the features and is notably the second application of the attention mechanism. This dual-attention mechanism strategy is what the authors aptly refer to as a \emph{rendezvous}. The rendezvous attention mechanism employs self-attention and cross-attention to capture complex semantic features in the instrument, verb, and target features.
In contrast to the volume-based predictions in \citet{1nywoye_rec_action_triplets}, RDV opts for a simpler classifier for prediction tasks. To fine-tune the model and optimize its performance, RDV employs uncertainty weighting \citep{uncertainty_weighting} to automatically determine the hyperparameters for each loss function.

\begin{table}[htb!]
\caption{Results of methods on the CholecT45 dataset. Results should be compared with caution as each reported work sometimes reported a different training/testing split. The mean of the published results are reported.} \label{triplet_results:table}
\resizebox{\columnwidth}{!}{
\begin{tabular}{|l|l|l|l|l|l|l|}
\hline
 & \multicolumn{3}{|c|}{\textbf{Component average precision}} & \multicolumn{3}{|c|}{\textbf{Triplet average precision}}  \\
\hline
 Method & $AP_{I}$ & $AP_{V}$ & $AP_{T}$ & $AP_{IV}$ & $AP_{IT}$ & $AP_{IVT}$ \\
\hline
\citet{1nywoye_rec_action_triplets} & 89.9 & 59.9 & 37.4  & 31.8 & 27.1 & 24.4 \\
\citet{NWOYE2022102433_rendevous} & 89.3 & 62.0 & 40.0 & 34.0 & 30.8 &  29.4 \\
\citet{sharma2022RendezvousInTime} & 88.6 & 64.0 & 43.4 & 38.3 & 36.9 &  29.7 \\
\citet{multi_self_distillation} & - & - & - & - & - &  36.1 \\
\hline
\end{tabular}
}
\end{table}

The third instalment in this series of papers, \citet{sharma2022RendezvousInTime} called Rendezvous in Time (RIT), follows the preceding works. While RDV focuses solely on single-frame features for triplet recognition, RIT incorporates temporal modelling into the RDV model.
In particular, \citet{sharma2022RendezvousInTime} introduce the Class Activation Guided Temporal Attention Module (CAGATAM), designed to enhance verb prediction and build upon the foundation of CAGAM. The Temporal Attention Module (TAM) plays a role in the temporal fusion of verb features extracted from the current and past frames, weighted by attention scores.

\citet{multi_self_distillation} apply the principle of self-distillation to solve the problem of class imbalance and label ambiguity in surgical action triplet recognition task. The method utilizes MTL and ensemble models as regularization to improve performance. A single teacher model with a Swin Transformer \citep{liu2021Swin} backbone and a classifier are trained on hard labels with binary cross entropy loss for the three tasks of instrument, action and target detection separately. Then, the Sigmoid probabilities from the teacher model are used to train three self-distilled student models as an ensemble to minimize a distillation loss.      

In the latest addition to the RDV series of papers, \citet{sharma2023surgical} introduce the surgical action triplet detection task: the localization of surgical instruments along with the recognition of surgical action triplets. 
The authors address a challenge in datasets designed for binary triplet label recognition when used for detection tasks.
Specifically, there may be multiple instruments in an image from the CholecT50 dataset~\citep{NWOYE2022102433_rendevous}, but only the primary instrument that is most prominent is annotated. This leads to a data association problem. 
To tackle this issue, the authors propose a two-stage network named the Multi-class Instrument-aware Transformer - Interaction Graphs (MCIT-IG). The first stage is referred to as a multi-class instrument-aware Transformer (MCIT) that performs target prediction sub-task by using the knowledge of the current instruments and their classes. It does this by first detecting all the instruments in an image using a deformable DETR model \citep{zhu2021deformable} trained on an annotated Cholec80 dataset, then extracting global features from the same image by using a feature extractor chained with a small transformer to provide global features, and finally combining the instrument detection information (from DETR) and image global feature information (from feature extractor) with a lightweight transformer to predict targets for each instrument. 
The second stage (IG) utilizes the instrument and target features from the first stage to construct a bipartite graph with action relationships as interaction edges. Although the supervision for the second stage is weak, a heuristic is employed to address the data association problem.
The authors note that the better the instrument localization prediction, the better the accuracy score for surgical action triplet prediction is. \newline

\indent In their work, \citet{chen2023surgical} point out the challenges of jointly optimizing three distinct classification problems as a single multiclass multilabel problem. They highlight that this problem is unbalanced, as positive results are only achieved when all three components of the triplet are predicted correctly. Moreover, the presence of multiple instruments in a single image and the lack of annotations for some instruments further introduce ambiguities to the triplet recognition task.
To address the complexities of triplet recognition, the authors propose a solution involving five smaller subnetworks. The first subnetwork is responsible for counting the number of tools and predicting the presence of key triplets or irrelevant triplets, including null actions and null targets.
The second subnetwork predicts the tool classes in the image, and similar to the Tripnet approach~\citep{1nywoye_rec_action_triplets}, it utilizes class activation maps. However, in this case, the Inflated 3D Convolutional Network (I3D) \citep{carreira2017quo} is used, which allows for the generation of class activation maps, which the authors found to be more accurate.
The third and fourth subnetworks jointly predict verbs and targets. Finally, the fifth subnetwork serves as both a fine-tuning and masking network, removing impossible triplets. It takes the logits predicted by the previous subnetworks and the current video clip as input to predict fine-tuned triplet logits and perform classification.
The authors emphasize that they train each subnetwork in different stages, as attempting to address multiple auxiliary tasks simultaneously can lead to task distraction and negative transfer. 

The surgical action triplet task is evolving in complexity. Initially framed as a multilabel multiclassification task for \textlangle{}instrument, verb, target\textrangle{} by \citet{1nywoye_rec_action_triplets, NWOYE2022102433_rendevous}, it now includes bounding box detection and segmentation of instruments \citep{sharma2023surgical} and video-based analysis \citep{1nywoye_rec_action_triplets}. The multilabel multiclassification task and its progression is similar to fields such as human-object interaction \citep{kim2021hotr} and panoptic scene generation \citep{yang2022panoptic} in computer vision, reflecting an increasing emphasis on spatial tasks and comprehensive scene understanding.

\subsubsection{Multi-granularity activity detection}
A comprehensive understanding of temporal relationships in MIS hinges on a network's ability to grasp the concept that multiple actions are required to complete a step and that multiple steps collectively constitute phases. This interconnection of granularities in surgical activities can serve as valuable signals for training deep neural networks in surgical activity recognition.
Despite the limited available datasets, there is a growing interest in multi-granularity surgical activity learning, spurred by the recent introduction of multiple datasets providing labels for various granularities \citep{autolaparo,huaulme2021micro,holistic_scene_understanding}. 

\begin{figure}[t]
   \centering
  \includegraphics[width=0.48\textwidth]{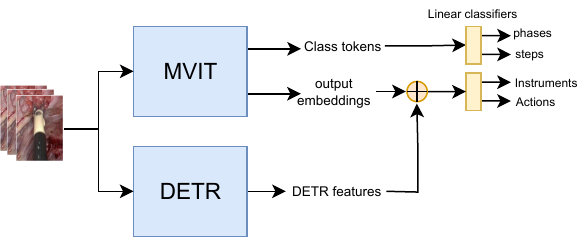}
  \caption{The TAPIR architecture from \citet{holistic_scene_understanding} for the prediction of phases, steps, instruments, and actions. It utilizes two separate encoders, a video feature extractor (MViT), and an instrument detector (DETR), along with various classification heads.}  
  \label{fig:tapir}
\end{figure}

In an early endeavour to simultaneously predict phase and step activities, \citet{Ramesh2021_phase_step_gastric_bypass} offer an innovative approach. Their proposed network adopts a two-stage design similar to the architecture of  EndoNet \citep{EndoNet}. The initial stage comprises a feature extractor with a classifier, where two distinct linear classifier heads are responsible for predicting phase and step for $N$ frames. In the subsequent stage, the vectors extracted in the first stage for multiple consecutive frames are concatenated and injected into a temporal convolutional neural network for temporal modelling, which leads to improved predictions for both phase and step activities. This work introduced the Bypass40 \citep{Ramesh2021_phase_step_gastric_bypass} dataset and evaluated it using separate classification metrics for phase and step prediction, demonstrating improvements over off-the-shelf models. The authors experimentally show that joint training of phase and step recognition is beneficial.

\begin{table*}[htb!]
\caption{ Methods that utilize multitask learning for anticipation in surgical workflow optimization. *HP - hard parameter sharing, GAN - generative adverserial network  } \label{workflow_optimization_tasks:table}
\resizebox{\textwidth}{!}{
\begin{tabular}{|l|l|l|l|l|l|}
\hline
  & Publication & Tasks & Optimization & Architecture & Speed(fps)\\
\hline

\multirow{3}{*}{ \parbox{3.0cm}{Anticipation in surgical workflow optimization.}}  & \citet{twinanda2018rsdnet} & \thead[l]{remaining surgery duration\\progress estimation} & linear scalarization & multistage network (HP) & - \\
\bluecline{2-6}

 & \citet{rivoir2020rethinking} & \thead[l]{instrument anticipation time \\anticipated instrument state} & linear scalarization & multistage network (HP) & -\\
\bluecline{2-6}

 & \citet{yuan2022anticipation} & \thead[l]{instrument anticipation time\\ phase anticipation time} & linear scalarization & multistage network (HP) & 34 \\
\bluecline{2-6}

 & \citet{ban2022supr} & \thead[l]{phase anticipation \\  phase recognition} & linear scalarization  & GAN network (HP) & -  \\

\bluecline{2-6}
 & \citet{jin2022trans} & \thead[l]{phase anticipation \\  phase recognition} & linear scalarization  & multistage network (HP) & 74.63\\
 
\hline

\end{tabular}
}
\end{table*}

The authors of \citet{holistic_scene_understanding} aim to completely address the temporal scene understanding problem in MIS.
They introduce a multi-level activity detection model illustrated in \figref{fig:tapir} named Transformer for Action, Phase, Instrument, and Steps Recognition (TAPIR) to achieve this goal.
TAPIR leverages two distinct backbones to capture various types of information. The first, the Multiscale Vision Transformer (MViT) \citep{fan2021multiscale}, for extracting global and temporal features from sequential video frames. The second, Deformable Transformers for End-to-End Object Detection (Deformable DETR), focuses on capturing spatial features relevant to instrument detection and box proposals.
The authors use a concatenation approach to combine the insights from these two backbones. The concatenated features from the MVIT and DETR are passed to linear classifiers to predict the action and instrument detection tasks. As for the prediction of phases and steps, this is solely carried out using the class tokens from the MViT backbone and linear classifiers. This work introduced the PSI-AVA dataset and evaluated it using mean average precision for phase, steps, actions, and instrument detection. The authors demonstrate improvements over off-the-shelf models and show that multitask frameworks outperform their single-task counterparts.

\citet{holistic_scene_understanding,Ramesh2021_phase_step_gastric_bypass} demonstrate the benefits of jointly learning multiple activities. Further research investigating the relationships between different surgical workflow granularities using the provided datasets would be valuable to the community. Additionally, metrics that measure the joint prediction of multiple granularities together, thereby assessing the accuracy of relationship prediction between granularities, would also be useful.

\subsection{MTL for anticipation in surgical workflow optimization}
Optimizing surgical workflows is critical for enhancing the efficiency and safety of procedures in the Operating Room (OR) \cite{yuan2022anticipation}. Multitask learning (MTL) has been employed to improve various aspects of surgical workflow, such as predicting future instrument usage, phase transitions or time till the end of the ongoing surgery. This subsection reviews four studies that utilize MTL for predicting future instruments and phases to optimize surgical workflows in MIS.
\tabref{workflow_optimization_tasks:table} summarizes the papers discussed in this subsection, highlighting the multiple tasks addressed, the optimization approaches adopted, the architectural strategies used for multitask learning, and the reported speeds for each study.

\citet{twinanda2018rsdnet} propose RSDNet, a deep learning pipeline to estimate remaining surgery duration (RSD) and surgery progress using visual information from laparoscopic videos. They employ a two-stage network with a ResNet feature extractor and an LSTM to predict the remaining time until the end of the surgery and the percentage if surgery completed. The authors demonstrate that their approach outperforms naive strategies.

\citet{rivoir2020rethinking} introduce a method to anticipate surgical instrument usage with sparse annotations using Bayesian deep learning \cite{NIPS2017_2650d608}. They employ a Bayesian AlexNet-style network \cite{alexnet} and a Bayesian LSTM \cite{LSTM} to handle uncertainty and predict the remaining time until an instrument is needed (regression) and the current state of an instrument (whether it will be used soon, is being used, or is not needed). Experiments on the Cholec80 dataset \cite{EndoNet} reveal that prediction uncertainty varies by instrument, with some instruments strongly related to specific phases showing low uncertainty, while others are more difficult to predict.

\citet{yuan2022anticipation} propose the Instrument Interaction Aware Anticipation Network (IIA-Net) to predict surgical phases and instrument usage. IIA-Net consists of two stages: a spatial feature extractor and a temporal model. The spatial feature extractor, a multitask model with a ResNet50 backbone, extracts visual features, predicts tools and current phase, and captures geometric and semantic features of instrument interactions through an Instrument Interaction Module. The temporal stage is a causal dilated multi-stage temporal convolutional network (MS-TCN) for temporal pattern recognition, which uses predictions from the first stage to anticipate phases and instrument usage.

\citet{ban2022supr} and \citet{jin2022trans} use multitask learning to perform workflow recognition as well as workflow anticipation.\citet{ban2022supr} present SUPR-GAN, a generative adversarial network (GAN) designed to predict future surgical phase trajectories in laparoscopic surgery. SUPR-GAN generates possible future phase trajectories with a generator and uses a discriminator to ensure phase accuracy. The generator, a CNN-LSTM with an encoder-decoder architecture, predicts future phase trajectories based on past video frames, while the discriminator, an LSTM, distinguishes between real and fake trajectories.
Although primarily focused on future phase prediction, SUPR-GAN employs a multitask learning framework by integrating current phase recognition and future phase prediction tasks. This multitask approach allows the model to utilize shared temporal features, enhancing predictive accuracy and robustness. The authors use per-transition accuracy to evaluate phase transitions and Levenshtein distance \citep{levenshtein1966binary} to measure the similarity between predicted phase sequences and the ground truth.

\begin{table*}[htb!]
\caption{Methods that utilize multitask learning for skill assessment. * HP - hard parameter sharing. } \label{skills_tasks:table}
\resizebox{\textwidth}{!}{
\begin{tabular}{|l|l|l|l|l|l|}
\hline
  & Publication & Tasks & Optimization & Architecture & Speed(fps)\\
\hline

\multirow{2}{*}{ \parbox{3.0cm}{skill assessment}}  & \citet{jian2020multitask} & \thead[l]{surgical skill level classification\\surgical skill score prediction} & linear scalarization &  shared encoder with multiple  task branches (HP) & - \\
\bluecline{2-6}

 & \citet{accurate_skill_assessment_gesture} & \thead[l]{surgical gesture recognition \\ surgical skill level classification(global) \\ surgical skill score prediction(global) \\  skill score prediction(intermediate)\\ gesture evaluation(intermediate)} & uncertainty weighting & multistage network (HP) & - \\

\hline

\end{tabular}
}
\end{table*}

 \citet{jin2022trans} presents Trans-SVNet, a novel hybrid embedding aggregation model using Transformers and multitask learning to address surgical workflow analysis. It leverages both spatial and temporal embeddings to improve the performance of workflow recognition and workflow anticipation. The paper introduces a Transformer-based model that combines spatial and temporal embeddings, enabling better preservation of spatial details and improved temporal information extraction.The authors experimentally show that their model improves performance in both tasks on multiple surgical video datasets.

These  five methods \citet{twinanda2018rsdnet,rivoir2020rethinking,yuan2022anticipation,ban2022supr,jin2022trans} highlight several key benefits of multitask learning, such as enhanced predictive accuracy through shared representation, robustness against noise and sparsity \cite{rivoir2020rethinking,jin2022trans}, and real-time inference capabilities without compromising accuracy \cite{ban2022supr,yuan2022anticipation}. Despite their methodological differences, all five studies underscore the importance of integrating multiple tasks to improve predictive performance and robustness.
While phase anticipation is valuable, it would also be beneficial to test these phase anticipation techniques on other public datasets, particularly those with finer granularity than phases. For instance, the PSI-AVA dataset \citep{holistic_scene_understanding} with 21 steps or the MultiBypass140 dataset \citep{Lavanchy2024} with 46 steps could provide further insights.

\subsection{MTL for surgical skill assessment}

Traditionally, surgical skill assessment involves senior surgeons observing and evaluating less experienced counterparts using standardized rating checklists such as the Objective Structured Assessment of Technical Skills (OSATS) \citep{osats}. However, the demand for training more doctors exceeds the number of experienced surgeons, making an automated system for scoring a surgeon's skill invaluable for trainees.
In this subsection, we discuss works that utilized multitask learning for surgical skill assessment. The JHU-ISI Gesture and Skill Assessment Working Set (JIGSAWS) dataset \citep{jigsaws} is the standard benchmark for training and evaluating surgical gestures and skills \cite{lam2022machine} in minimally invasive surgeries. Common metrics used in these evaluations are accuracy for gesture recognition and Spearman's correlation ($\rho_{GRS}$ and $\rho_{OSATS}$) for skill scores.
\tabref{skills_tasks:table} summarizes the papers discussed in this subsection, highlighting the multiple tasks addressed, the optimization approaches adopted, the architectural strategies used for multitask learning, and the reported speeds for each study.

In \citet{jian2020multitask}, the authors introduce a multitask network designed to tackle the dual objectives of overall surgical skill level classification (categorizing surgeons as expert, intermediate, or novice) and attribute score regression (modified OSATS attributes from datasets like JIGSAWS \citep{jigsaws}). This network architecture revolves around a shared encoder featuring two heads: one for classification and the other for the regression task. The core of this encoder is a variation of  I3D \citep{carreira2017quo}, a proven feature extractor in the domain of action recognition. To enhance the I3D capabilities, the encoder incorporates attention modules that enable it to focus on crucial segments of the video clips.
To facilitate efficient computation, the authors split the input videos into $K$ equal parts, and clip snippets are sampled within these segments.  Subsequently, a classification head is employed to predict the overall surgical skill assessment score, while a regression head predicts individual attribute scores. The classification and regression scores for each snippet are then aggregated, with the average of output features forming the final solution.

\begin{table*}[ht!]
\caption{Methods that utilize multitask learning for surgical report generation. *HP - hard parameter sharing.   } \label{report:table}
\resizebox{\textwidth}{!}{
\begin{tabular}{|l|l|l|l|l|l|}
\hline
  & Publication & Tasks & Optimization & Architecture & Speed(fps)\\
\hline

\multirow{2}{*}{ \parbox{3.0cm}{report generation}}  & \citet{seenivasan2022global} & \thead[l]{tool segmentation\\tool-tissue interaction} & sequential training &  shared encoder with interacting task branches (HP) & - \\
\bluecline{2-6}

 & \citet{mobarakol_task_aware} & \thead[l]{tool-tissue interaction \\ scene caption optimization} & sequential training &  shared encoder with multiple task branches  (HP) & - \\

\hline

\end{tabular}
}
\end{table*}

\begin{figure*}[htb!]
   \centering
  \includegraphics[width=0.9\textwidth]{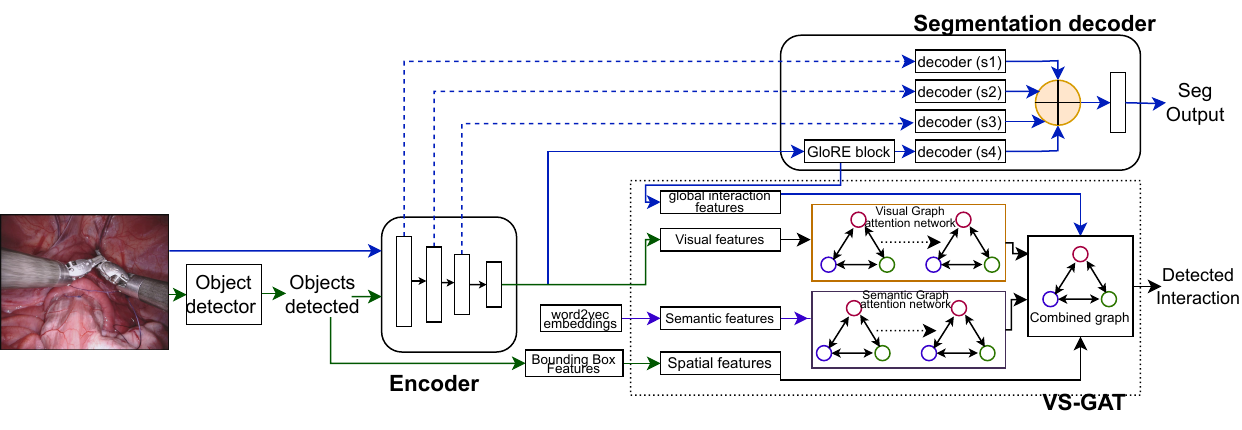}
  \caption{The proposed architecture for globally-reasoned multi-task surgical scene understanding in \citet{seenivasan2022global} for performing instrument segmentation and tool-tissue interaction. The model features a feature encoder, global and local reasoning for instrument segmentation and a VS-GAT \citep{9739429} model for interaction detection.} 
  \label{fig:global_report}
\end{figure*}

In \citet{accurate_skill_assessment_gesture}, the focus is on achieving interpretable skill assessment inspired by how senior clinicians assess other surgeons' performance. The authors observe that senior clinicians assess how well each surgical gesture is executed. The authors propose a two-stage network to replicate this assessment method.
The first stage primarily aims to detect surgical gestures in different clips along with predicting the skill level classification and skill level score as auxiliary tasks. The surgical video input is initially split into $C$ clips, designed to streamline the computational process. These clips are then processed by a shared C3D encoder adapted from \citet{tran2015learning}. The features obtained from all $C$ clips are concatenated and used as the input to the decoder.
The surgical gesture recognition decoder adopts a multi-stage temporal convolutional network. In parallel, the skill level classification and skill level prediction are executed through a shared LSTM, with classification and regression heads.
The second stage focuses on the task of predicting skill assessment for the gestures identified in the first stage. To achieve this, the predicted gestures from the initial stage serve as a basis for segmenting the original surgical video into distinct gesture clips. These clips are then utilized as input for a C3D network, initialized with the weights from the C3D encoder employed in the first stage. Similarly, the decoder for predicting gesture level immediate skill score is implemented using an LSTM, featuring a dedicated head for gesture level immediate skill score prediction. Loss weighting with uncertainties \citet{uncertainty_weighting} was utilized for multitask optimization. On the dataset, the authors extend the JIGSAWS dataset by having surgeons annotate each gesture with a binary label of good(1) or bad(0). 

Both \citet{accurate_skill_assessment_gesture} and \citet{jian2020multitask} demonstrate improvements over single-task surgical skill assessment models. However, the JIGSAWS dataset, released in 2014, remains the standard benchmark for surgical skill assessment \cite{lam2022machine}. Given its relatively small size compared to contemporary deep learning datasets, there is a need for larger and more comprehensive datasets in the field.

\subsection{MTL for surgical report generation} \label{sec::report_generation}

The automatic generation of surgical reports can free surgeons and nurses from the tedious task of document entry, allowing them to focus more on patients and post-operative interventions \citep{xu2021class,lin2022sgt}. 
Currently, in the existing literature, the approach to the surgical report generation task primarily revolves around frame-by-frame scene captioning \citep{xu2021class,lin2022sgt,seenivasan2022global,mobarakol_task_aware}.
The increasing interest in surgical report generation is closely tied to advancements in scene captioning and scene graph generation within the broader context of scene captioning research \citep{9739429,Cornia_2020_CVPR}
Notably, MTL has been applied to this field, incorporating aspects like image captioning and scene graph generation alongside other tasks to further enhance the overall capabilities of the system. The benchmark dataset, an extension of the EndoVis2018 segmentation dataset to include tool-tissue interactions, is used to form captions via sentence templates \cite{islam2020learning}.  We term this dataset EndoVis2018-with-interactions. The metrics used include tool-tissue interaction accuracy (ACC), mean average precision (mAP), recall (RE), and other metrics depending on the predicted tasks.

\tabref{report:table} summarizes the papers discussed in this subsection, highlighting the multiple tasks addressed, the optimization approaches adopted, the architectural strategies used for multitask learning, and the reported speeds for each study.

In \citet{seenivasan2022global}, a new approach is introduced for optimizing scene graphs to facilitate object-to-object interactions, focusing on the surgical report generation automation task. The authors frame this challenge as a multitask problem involving two core tasks: instrument segmentation and tool-tissue interaction. The authors propose a network with a single encoder and two decoders, as visualized in \figref{fig:global_report}.
The process commences with a preprocessing step, which generates bounding boxes and their corresponding classifications. These bounding boxes serve as a foundation for relationship modelling. The encoder, based on ResNet18, is responsible for shared feature extraction.
The first decoder concentrates on local and global reasoning for instrument segmentation, configured as a U-net-like decoder with two noteworthy differences. The first is that the outputs of each lower-resolution decoder block do not feed into the next higher-resolution. Instead, each decoder block produces an output directly, which is concatenated together and passed through a conv block to produce the final segmentation prediction.  Additionally, a GloRE block \citep{Chen_2019_CVPR} is integrated into the output features of the encoder, enriching the global reasoning capabilities in the feature space. The features obtained from the GloRE block, known as Global Reasoned features, are also leveraged in the tool-tissue interaction decoder.
The tool-tissue interaction decoder is structured as a scene graph from which the surgical report is generated. The authors employ the visual-semantic graph attention network (VS-GAT) \citep{9739429}. This network essentially comprises two components: a visual graph attention network and a semantic graph attention network. Both networks are harmoniously combined to create a unified graph. Notably, the authors introduce global interaction features from the GloRE block into this combined graph before instrument-tissue interaction prediction. The multitask optimization is a variation of the ideas of sequential training from \citet{learning_look_tracking}. 

\begin{table*}[ht!]
\caption{Methods that utilize large models for multiple tasks. *PE - post encoding i.e. speed assuming image encodings provided.   } \label{large_models:table}
\resizebox{\textwidth}{!}{
\begin{tabular}{|l|l|l|l|l|l|}
\hline
  & Publication & Tasks & Optimization & Architecture & Speed(fps)\\
\hline

\multirow{2}{*}{ \parbox{3.0cm}{Visual Question Answering}}  & \citet{seenivasan2022surgical} & VQA &  single loss  & modified VisualBERT & - \\
\bluecline{2-6}

 & \citet{seenivasan2023surgicalgpt} & VQA &  single loss   & modified GPT-2  & - \\
\bluecline{2-6}

\hline

\multirow{5}{*}{ \parbox{3.0cm}{Promptable Segmentation}}  & \citet{zhou2023text} & 
\thead[l]{promptable segmentation\\image reconstruction}
 & linear scalarization &  modified CLIP & 22(PE) \\
\bluecline{2-6}

 & \citet{wang2023sam} & promptable segmentation & zero-shot inference & SAM & 20(PE) \\
\bluecline{2-6}

 & \citet{ma2023segment} & promptable segmentation &  combined loss & SAM & - \\
\bluecline{2-6}

 & \citet{yue2023surgicalsam} & promptable segmentation &  combined loss & modified SAM & 2 \\
\bluecline{2-6}

 & \citet{paranjape2023adaptivesam} & promptable segmentation &  combined loss & modified SAM & -\\

\hline
\end{tabular}
}
\end{table*}

\citet{mobarakol_task_aware} explore surgical report generation as a two-task problem of scene graph optimization for object-to-object relationships in a scene and frame-by-frame image captioning. The network architecture developed for this purpose is built around a shared encoder and two decoders.
Based on the standard ResNet-18, the shared encoder forms the foundational feature extraction component. The object-to-object scene graph optimization decoder leverages the visual-semantic graph attention network from \citet{9739429}. This aspect of the network focuses on enhancing the understanding of complex relationships between objects within the surgical scene.
In parallel, the frame-by-frame image captioning task is addressed through the utilization of a meshed-memory transformer from \citet{Cornia_2020_CVPR}. 
The optimization process is facilitated through a sequential training method similar to  ATO \citep{islam2020apmtl}. 
Recognizing the challenge of model generalization to target domains and the need to accommodate factors like new instruments, the authors introduce a loss function inspired by continual learning, termed class incremental contrastive loss. 

The surgical reporting task is still in its early stages. While the EndoVis2018-with-interactions dataset, which incorporates sentence templates and classification questions, has been useful, it may not fully capture the complexity of natural language used in real-world surgical reports. Consequently, the relevance of captioning metrics and tasks based on this dataset can be questioned. Developing a dataset based on actual surgical reports generated by surgeons in day-to-day procedures would likely provide more valuable insights and improve the robustness of surgical report generation systems.

\subsection{Large models for solving multiple tasks}
Large models pretrained on extensive datasets, such as GPT-2 \cite{radford2021learning} and SAM \cite{kirillov2023segment}, have demonstrated remarkable generalization capabilities. They exhibit emergent properties, such as the ability to solve multiple tasks by utilizing task-promptable engineering \cite{radford2019language, kirillov2023segment}. Task-promptable engineering allows these models to adapt to various tasks based on given prompts, making them highly versatile and generalizable backbones capable of zero-shot learning or fine-tuning for different downstream applications \cite{su2024stem}. In the current literature, large models exhibit this capability of solving multiple tasks in surgical scene understanding, with fields such as task-promptable segmentation, and visual question answering being explored.

\tabref{large_models:table} summarizes the papers discussed in this subsection, highlighting the multiple tasks addressed, the optimization approaches adopted, the architectural strategies used for multitask learning, and the reported speeds for each study.

\subsubsection{Visual question answering}
\citet{seenivasan2022surgical} introduce the problem of visual question answering (VQA) in the context of surgery. They develop a large model capable of providing textual answers to a wide range of questions, including tasks such as tissue presence recognition, instrument localization, tool presence recognition, and action recognition. To achieve this, they expand the EndoVis2018 and Cholec80 datasets by adding sentence-based and classification-based answers to a predefined set of questions. The authors propose two models based on the VisualBERT architecture \citep{li2019visualbert} for classification-based and sentence-based answering in surgical VQA. They improve the VisualBERT encoder by introducing cross-token and cross-channel submodules to enhance the interaction between visual and text tokens. For the classification-based model, they utilize linear classifiers, with initial sentence classification determining the appropriate linear layer. The sentence-based model employs a standard transformer decoder.

In subsequent work, SurgicalGPT \citep{seenivasan2023surgicalgpt} for the surgical VQA task, the authors replace the VisualBERT architecture with GPT-2 \citep{radford2019language}, enhancing it with a vision encoder. They concatenate tokens from the text and vision encoders before feeding them to the GPT decoder, which they refer to as LV-GPT. Additionally, they change the unidirectional attention to a bidirectional attention model in the GPT decoder. Notably, they report improved performance compared to \citet{seenivasan2022surgical}. 

\citet{seenivasan2022surgical,seenivasan2023surgicalgpt} generated multiple VQA datasets by converting existing interaction-based minimally invasive surgery datasets, such as Endovis-with-interaction, Cholec80, and PSI-AVA, into Visual Question Answering (VQA) datasets. They demonstrated improvements over state-of-the-art VQA models for medical applications using these datasets.
Models trained on classification-based VQA datasets perform tasks such as counting, interaction recognition, localization, and classification by utilizing text prompts. However, it would be interesting to see these models applied to more complex VQA datasets with more complex answers, that require a really deep understanding of the surgical scene.  

\subsubsection{Promptable segmentation}
Another application of the concept of generalizable tasks is \emph{promptable segmentation}, which primarily focuses on binary segmentation for ideas represented in various forms, including text \citep{zhou2023text}, points \citep{kirillov2023segment}, bounding boxes \citep{kirillov2023segment}, and reference images \citep{lueddecke22_cvpr}. This approach allows for the execution of binary segmentation, instance segmentation, part segmentation, and instrument-type segmentation, provided that suitable prompts are provided.

An example of promptable segmentation in surgical scene understanding can be found in the work of \citet{zhou2023text}. Their work addresses the challenge of distinguishing and segmenting diverse surgical instruments using textual prompts. To achieve this, they leverage pretrained CLIP text and vision encoder foundational models and introduce a novel text promptable mask decoder. The authors report capabilities to generalize text promptable segmentation to tissue segmentation, instrument-part, and instrument-type segmentation. Notably, their method demonstrates strong generalization capabilities,  as evidenced by robust performance in cross-dataset evaluations.

Recent advancements in promptable segmentation have seen remarkable progress through methods based on the Segment Anything Model (SAM) \citep{kirillov2023segment}. SAM serves as a foundational model that accommodates different prompting strategies, namely: points, bounding box, segmentation mask, and text. 

SAM has demonstrated exceptional generalization abilities and has found applications in surgical segmentation tasks. An empirical study conducted in \citet{wang2023sam} examines the robustness and generalizability of SAM when applied to the EndoVis2017 and EndoVis2018 datasets. Their findings reveal that pretrained SAM excels in terms of generalizability, especially when used with bounding box prompts, achieving state-of-the-art results. However, it is important to acknowledge that comparing bounding box prompts to class-based segmentation techniques might not be entirely fair. 

\citet{wang2023sam} also observe that SAM does not perform well for surgical instrument segmentation with point-based prompts, and its generalizability can be compromised under conditions of data corruption commonly encountered in surgical segmentation tasks. In summary, while SAM exhibits strong generalization capabilities, it relies on users to supply real-time bounding boxes for each image. In addition, SAM may not be robust in the presence of noise and data corruption. To address these challenges, several innovative approaches have been proposed.

MEDSAM \citep{ma2023segment} is a foundational model designed for universal medical image segmentation, curated from a dataset comprising over one million images, including CT scans, histology images, and surgical scenes. This model aims to bridge the gap between SAM for natural images and SAM for medical images. It adopts bounding box prompts as input, with a key distinction being the freezing of the SAM encoders, while only training the SAM decoder.

SurgicalSAM \citep{yue2023surgicalsam} proposes using the concepts of a prototype image, a real image of a particular instrument that captures the interested image, as a prompt instead of bounding boxes per image as prompts or directly using class names. More specifically, their method builds a memory bank of prototype images then they use the prototype-based class prompt encoder to exploit similarities between images in the dataset and class prototype images to create prompts, which they call prompt embeddings. In addition, the authors also propose a contrastive prototype learning loss to ensure that during training, the feature space for each different prototype is far from each other. 

AdaptiveSAM \citep{paranjape2023adaptivesam} addresses the bounding box requirement by utilizing text as prompts. It employs text embeddings from pretrained CLIP and passes them through a trainable affine layer before applying them to the SAM prompt decoder. A notable feature of AdaptiveSAM is the introduction of \emph{bias tuning} as a more memory-efficient method to adapt the SAM encoders. This approach trains the bias of the multi-head attention layers in the SAM image encoder and the normalization layers, achieving adaptation with higher efficiency.

All these methods \citep{ma2023segment,yue2023surgicalsam,paranjape2023adaptivesam} report significant improvement over the vanilla SAM approach when applied to surgical segmentation datasets over various segmentation tasks.

\section{Public datasets for MTL in MIS} \label{sec::datasets}

\begin{table*}[htb!]
\caption{A summary of publically available multitask learning datasets for minimally invasive surgeries with information on the dataset characteristics and annotation characteristics. Datasets annotated for a single task that may have been used in MTL research such as Endovis2017, Endovis2018 and Cholec80 are omitted from the table for clarity.} \label{table:dataset_summary}
\resizebox{\textwidth}{!}{  
\begin{tabular}{|l|l|l|l|l|l|l|}
  \thickhline
  Dataset   & Brief description & Procedure & 
  Task & Input & Annotations & Paper  \\

  \thickhline
  \multirow{4}{*}{\parbox{1.6cm}{
  \href{https://autolaparo.github.io/}{AutoLaparo}
  }} & \multirow{4}{*}{\parbox{9.5cm}{A dataset for image-guided surgical automation in laparoscopic hysterectomy comprising three sub-datasets: surgical workflow recognition, laparoscope motion prediction, instrument and anatomy segmentation. } }
  &
  \multirow{4}{*}{\parbox{0.1cm}{21} }
  & Phase recognition task
  & 1388 minutes of surgical videos  
  & 1388 minutes of phase labels 
  & \multirow{2}{*}{\parbox{2.0cm}{\citep{autolaparo} }}  \\  \bluecline{4-6}
  \arrayrulecolor{black}
  & &
  & Laparoscope motion prediction
  & 300 video clips   
  & 300 next motion mode labels
  &
  \\ \bluecline{4-6}
  \arrayrulecolor{black}
  & & 
  & Instrument part segmentation
  & 1800 keyframes
  & 1800 part segmentation maps
  &
  \\ \bluecline{4-6}
  \arrayrulecolor{black}
  & &  
  & Key anatomy segmentation
  & 1800 keyframes
  & 1800 tissue segmentation maps
  & 
  \\ 

  \thickhline
  \multirow{5}{*}{\parbox{1.6cm}{   \href{https://www.synapse.org/\#!Synapse:syn25101790/wiki/}{HeiSURF}
  }} 
  & \multirow{5}{*}{\parbox{9.5cm}{ A laparoscopic cholecystectomy designed for surgical activity recognition, full scene segmentation, and skill assessment. It is a parent dataset to the HeiChole dataset} }
  & \multirow{5}{*}{\parbox{0.1cm}{33} }
  & Phase recognition
  & 23.3 hours of surgical videos
  & 23.3 hours of phase labels
  & \multirow{2}{*}{\parbox{2.0cm}{\citep{heisurf}} }  \\  \bluecline{4-6}
  & &  
  & Full scene segmentation 
  & 827 keyframes 
  & 827 full scene segmentation maps
  &
  \\ \bluecline{4-6}
  & &  
  & Action recognition 
  & 5514 instances in frames 
  & 5514 action labels
  &
  \\ \bluecline{4-6}
  & &  
  & Tool presence prediction
  & 6980 instances in frames 
  & 6980 tool presence labels
  &
  \\ \bluecline{4-6}
  & &  
  & Surgical skill-score prediction
  & 99 video clips 
  & 495 skill scores
  &  
  \\ 

  \thickhline
  \multirow{4}{*}{\parbox{1.6cm}{
  \href{https://github.com/BCV-Uniandes/TAPIR}{PSI-AVA}
  }} 
  & \multirow{4}{*}{\parbox{9.5cm}{ A dataset of robot-assisted radical prostatectomy designed for research into the complementary nature of surgical activity recognition tasks.} }
  & \multirow{4}{*}{\parbox{0.1cm}{ 8 } }
  & Phase recognition
  & 20.45 hours of surgical videos
  & 20.45 hours of phase labels
  & \multirow{2}{*}{\parbox{2.0cm}{\citep{holistic_scene_understanding}} } \\  \bluecline{4-6}
  & &  
  & Step recognition
  & 20.45 hours of surgical videos
  & 20.45 hours of step labels
  &
  \\ \bluecline{4-6}
  & &  
  & Action recognition
  & 5804 instances in frames 
  & 5804 action labels
  &
  \\ \bluecline{4-6}
  & &  
  & Instrument detection
  & 5804 instances in frames 
  & 5804 bounding boxes with labels
  &
  \\ 

  \thickhline
  \multirow{2}{*}{\parbox{1.6cm}{  \href{https://www.synapse.org/\#!Synapse:syn21903917/wiki/601992}{HeiCo}
  }} & \multirow{2}{*}{\parbox{9.5cm}{ A dataset of three different colorectal procedures (proctocolectomy, rectal resection, and sigmoid resection procedures) with emphasis on dataset generalization and diversity.} }
  & \multirow{2}{*}{\parbox{0.1cm}{30} }
  & Phase recognition
  & 9.45 hours of surgical videos
  & 9.45 hours of phase labels
  & \multirow{2}{*}{\parbox{2.0cm}{\citep{heico}} }\\  \bluecline{4-6}
  & &  
  & Instrument instance segmentation
  & 10,040 keyframes 
  & 10,040 instance segmentation maps
  & 
  \\ 
  & &  
  & 
  & 
  & 
  & 
  \\

  \thickhline
  \multirow{5}{*}{\parbox{1.6cm}{\href{https://www.synapse.org/\#!Synapse:syn21776936/wiki/601700}{MISAW} 
  }} 
  & \multirow{5}{*}{\parbox{9.5cm}{ A micro-surgical anastomosis dataset with a focus on evaluating the impact of learning multiple surgical activity recognition tasks together.} }
  & \multirow{2}{*}{\parbox{0.1cm}{ 27 } }
  & Phase recognition
  & 1.5 hours of surgical videos
  & 1.5 hours of phase labels
  & \multirow{2}{*}{\parbox{2.0cm}{\citep{huaulme2021micro}} }\\  \bluecline{4-6}
  & &  
  & Step recognition
  & 1.5 hours of surgical videos
  & 1.5 hours of step labels
  &
  \\ \bluecline{4-6}
  & &  
  & Action recognition
  & 1.5 hours of surgical videos 
  & 1.5 hours of action labels
  &
  \\ \bluecline{4-6}
  & &  
  & Tool presence prediction
  & 1.5 hours of surgical videos
  & 1.5 hours of tool presence labels
  &
  \\ \bluecline{4-6}
  & &  
  & target tissue prediction
  & 1.5 hours of surgical videos
  & 1.5 hours of tool target labels
  &
  \\ 

  \thickhline
  \multirow{5}{*}{\parbox{1.6cm}{\href{https://www.synapse.org/\#!Synapse:syn25147789/wiki/}{PETRAW} 
  }} 
  & \multirow{5}{*}{\parbox{9.5cm}{A dataset containing the peg transfer task in laparoscopic surgery training. It is designed for multiple surgical activity recognition tasks. This dataset is collected from a virtual reality simulator} }
  & \multirow{2}{*}{\parbox{0.1cm}{ 150 } }
  & Phase recognition
  & 5.86 hours of surgical videos
  & 5.86 hours of phase labels
  & \multirow{2}{*}{\parbox{2.0cm}{\citep{huaulme2022peg}} } \\  \bluecline{4-6}
  & &  
  & Step recognition
  & 5.86 hours of surgical videos
  & 5.86 hours of step labels
  &
  \\ \bluecline{4-6}
  & &  
  & Action recognition
  & 5.86 hours of surgical videos 
  & 5.86 hours of action labels
  &
  \\ \bluecline{4-6}
  & &  
  & target tissue prediction
  & 5.86 hours of surgical videos
  & 5.86 hours of tool target labels
  &
  \\ \bluecline{4-6}
  & &  
  & Full scene segmentation
  & 5.86 hours of surgical videos
  & 5.86 hours of scene segmentation
  &
  \\ 

  \thickhline 
  \multirow{3}{*}{\parbox{1.6cm}{  \href{https://github.com/CAMMA-public/MultiBypass140}{Multi-Bypass140}
  }} 
  & \multirow{3}{*}{\parbox{9.5cm}{A phase and step recognition dataset, particularly focusing on datasets collected from different hospitals (multi-centric) and the importance of multi-centric datasets for generalization. } }
  & \multirow{3}{*}{\parbox{0.1cm}{140} }
  & Phase recognition
  & 214hrs of surgical videos
  & 214hrs of phase labels
  & \multirow{3}{*}{\parbox{2.0cm}{\citep{Lavanchy2024}} }  \\  \bluecline{4-6}
  & &  
  & Step recognition
  & 214hrs of videos
  & 214hrs of step labels
  &
  \\ 
  & &  
  & 
  &
  &
  &
  \\

  \thickhline
  \multirow{2}{*}{\parbox{1.6cm}{
  \href{https://www.synapse.org/\#!Synapse:syn27618412/wiki/616881}{SAR-RARP50}
  }} 
  & \multirow{2}{*}{\parbox{9.5cm}{A dataset of suturing segments of robotic assisted radical prostatectomy. The dataset focuses on tool segmentation and surgical action recognition.} }
  & \multirow{2}{*}{\parbox{0.1cm}{50} }
  & Phase recognition
  & 3 hours of surgical videos
  & 3 hours of action labels
  & \citep{psychogyios2023sar}\\  \bluecline{4-6}
  & &  
  & Instrument segmentation
  & 1000 keyframes
  & 1000 Instrument segmentation maps
  &
  \\ 
  & &  
  & 
  &
  &
  &
  \\ 

  \thickhline
  \multirow{3}{*}{\parbox{1.6cm}{
  \href{https://saras-mesad.grand-challenge.org/Home/}{SARAS-MESAD}
  }} 
  & \multirow{3}{*}{\parbox{9.5cm}{A dataset containing both real and virtual human prostatectomy procedures for research into surgical activity recognition and instrument detection with a focus on cross-domain learning} }
  & \multirow{3}{*}{\parbox{0.1cm}{9} }
  & Instrument detection
  & 59k frames
  & 59k frames with bounding boxes 
  & \multirow{3}{*}{\parbox{2.0cm}{\citep{saras_mesad}} }  \\  \bluecline{4-6}
  & &  
  & Action recognition
  & 59k frames
  & 59k frames with action labels
  &
  \\ 
  & &  
  & 
  &
  &
  &
  \\ 

  \thickhline
  \multirow{3}{*}{\parbox{1.6cm}{
  \href{https://cholectriplet2021.grand-challenge.org/}{CholecT50}
  }} 
  & \multirow{3}{*}{\parbox{9.5cm}{The CholecT50 is designed for fine-grained action recognition and tool-tissue interaction in laparoscopic cholecystectomy surgeries. CholecT50 is the latest in the Cholec series of datasets. } }
  & \multirow{4}{*}{\parbox{0.1cm}{ 50} }
  & Tool presence recognition  
  & 100.9k frames  
  & 100.9k frames with tool presence labels 
  & \multirow{2}{*}{\parbox{2.0cm}{\citep{NWOYE2022102433_rendevous}} } \\  \bluecline{4-6}
  & &  
  & Action recognition
  & 100.9k frames
  & 100.9k frames with action labels
  &
  \\ \bluecline{4-6}
  & &  
  & Tissue target recognition
  & 100.9k frames
  & 100.9k frames with target labels
  & 
  \\

  \thickhline
  \multirow{3}{*}{\parbox{1.6cm}{
  \href{https://github.com/kamruleee51/ART-Net}{ART-Net}
  }} & 
  \multirow{3}{*}{\parbox{9.5cm}{A non-robotic laparoscopic hysterectomy dataset designed for 3D graphics applications like 3D measurement and Augmented Reality~(AR). } }  
  & \multirow{3}{*}{\parbox{0.1cm}{29} }
  & Tool presence prediction
  & 1500 frames
  & 1500 frames with tool presence labels
  & \multirow{2}{*}{\parbox{2.0cm}{\citep{art_net}} }  \\  \bluecline{4-6}
  & &  
  & Binary instrument segmentation
  & 635 keyframes 
  & 635 binary instrument segmentation maps
  &
 \\ \bluecline{4-6}
  & &  
  & Geometric map prediction
  & 635 keyframes 
  & 635 geometric map labels
  &
 \\

  \thickhline 
  \multirow{3}{*}{\parbox{1.6cm}{  \href{https://github.com/lalithjets/SurgicalGPT}{EndoVis-18-VQA}
  }} 
  & \multirow{3}{*}{\parbox{9.5cm}{A dataset designed for advancing research in visual question answering using the Endovis2018 dataset. It contains classification-based questions about tissues, actions and tool locations. } }
  & \multirow{3}{*}{\parbox{0.1cm}{14} }
  & VQA
  & 11.7k questions
  & 2k frames  
  & \multirow{3}{*}{\parbox{2.0cm}{\citep{seenivasan2022surgical}} }  \\  
  & &  
  & 
  & 
  & 
  &
  \\ 
  & &  
  & 
  &
  &
  &
  \\
  \hline 

  \thickhline 
  \multirow{3}{*}{\parbox{1.6cm}{  \href{https://github.com/lalithjets/SurgicalGPT}{Cholec80-VQA}
  }} 
  & \multirow{3}{*}{\parbox{9.5cm}{A Surgical VQA dataset based on the Cholec80 dataset. It contains classification-based questions about surgical phases and instrument presence. } }
  & \multirow{3}{*}{\parbox{0.1cm}{40} }
  & VQA
  & 43k questions
  & 21.6k frames  
  & \multirow{3}{*}{\parbox{2.0cm}{\citep{seenivasan2022surgical}} }  \\  
  & &  
  & 
  & 
  & 
  &
  \\ 
  & &  
  & 
  &
  &
  &
  \\
  \hline  

  \thickhline 
  \multirow{3}{*}{\parbox{1.6cm}{  \href{https://github.com/lalithjets/SurgicalGPT}{PSI-AVA-VQA}
  }} 
  & \multirow{3}{*}{\parbox{9.5cm}{A Surgical VQA dataset based on the PSI-AVA dataset . It contains classification-based questions about the surgical phase, step, and location of surgical tools. } }
  & \multirow{3}{*}{\parbox{0.1cm}{8} }
  & VQA
  & 10.3k questions
  & 2.2k frames
  & \multirow{3}{*}{\parbox{2.0cm}{\citep{seenivasan2023surgicalgpt}} }  \\  
  & &  
  & 
  & 
  & 
  &
  \\ 
  & &  
  & 
  &
  &
  &
  \\
  \hline 

  \thickhline 
  \multirow{4}{*}{\parbox{1.6cm}{  \href{https://github.com/CAMMA-public/SSG-VQA}{SSG-VQA}
  }} 
  & \multirow{4}{*}{\parbox{9.5cm}{A Surgical VQA dataset comprising images from the CholecT45 dataset. It contains complex questions about tool location and presence, tissue location and presence, instrument action, instrument targets, object colours, and their relationships } }
  & \multirow{4}{*}{\parbox{0.1cm}{45} }
  & VQA
  & 960k questions
  & 25k frames 
  & \multirow{4}{*}{\parbox{2.0cm}{\citep{yuan2024advancing}} }  \\  
  & &  
  & 
  & 
  & 
  &
  \\ 
  & &  
  & 
  &
  &
  &
  \\
  & &  
  & 
  &
  &
  &
  \\
  \hline    

  \thickhline
  \multirow{2}{*}{\parbox{1.6cm}{\href{https://cirl.lcsr.jhu.edu/research/hmm/datasets/jigsaws_release/}{JIGSAWS}}} 
  & \multirow{3}{*}{\parbox{9.5cm}{An automatic gesture recognition and surgical skill assessment dataset containing videos of surgeons performing suturing, knot-tying and needle-passing on a bench-top model with a da Vinci. } }
  & \multirow{3}{*}{\parbox{0.1cm}{ 104 } }
  & Surgical gesture classification
  & 208 minutes of surgical videos
  & 208 minutes of surgical gesture labels  
  & \multirow{3}{*}{\parbox{2.0cm}{\citep{jigsaws}} }\\  \bluecline{4-6}
  & &  
  & Surgical skill-score prediction
  & 104 video clips 
  & 104 global surgical skill scores
  & 
  \\ 
  & &  
  & 
  & 
  & 
  & 
  \\   

  \thickhline 
  \multirow{3}{*}{\parbox{1.6cm}{  \href{https://github.com/CAMMA-public/Endoscapes}{Endoscapes}
  }} 
  & \multirow{3}{*}{\parbox{9.5cm}{Endoscapes dataset was designed for advancing research on the Critical View of Safety (CVS), segmentation and object in Laparoscopic Cholecystectomy surgeries.} }
  & \multirow{3}{*}{\parbox{0.1cm}{201} }
  & CVS prediction
  & 58.8k frames
  & 58.8k frames with CVS labels 
  & \multirow{3}{*}{\parbox{2.0cm}{\citep{murali2023endoscapes}} }  \\  \bluecline{4-6}
  & &  
  & Scene object detection
  & 11k frames
  & 11k frames with bounding boxes
  &
  \\  \bluecline{4-6}
  & &  
  & Scene instance segmentation
  & 422 frames
  & 422 frames with instance seg masks 
  &
  \\
  \hline

\end{tabular}
}
\end{table*}

In this section, we explore the public datasets curated to support MTL for MIS. These datasets offer a valuable foundation for researchers to experiment, innovate, and address real-world MIS challenges. 

Due to the inherent difficulties in producing surgical video datasets, the availability of public datasets suitable for MTL in this domain is currently limited. While there are some datasets designed specifically for MTL, such as the \textit{multi-granularity surgical activity recognition datasets}, the number of datasets catering to other applications of MTL in surgical vision remains scarce. 
Nevertheless, it is worth noting that some authors have successfully explored approaches to leverage single-task datasets for MTL by generating self-supervised auxiliary tasks from the existing data.
While we acknowledge that some single task datasets such as Endovis2017 \citep{endovis2017}, Endovis2018 \citep{endovis2018}, Cholec80 \citep{EndoNet}, and m2cai \citep{jin2018tool} have played a key role in some MTL research,
we do not detail these single task datasets in this section.  Moreover, there are other multitask datasets that have received less attention in the literature, which we aim to highlight. We present a comprehensive list of datasets designed to address multiple tasks in surgical vision.

For a quick overview of datasets designed for multiple tasks, along with information such as the amount of of images and labels, and other relevant information, readers can refer to \tabref{table:dataset_summary}.
We do not include private datasets (e.g. ByPass40 \citep{Ramesh2021_phase_step_gastric_bypass}). For more information about surgical tool datasets, including non-multitask learning datasets, we recommend the following online resource: \href{ https://github.com/luiscarlosgph/list-of-surgical-tool-datasets}{https://github.com/luiscarlosgph/list-of-surgical-tool-datasets}.   

\subsection{Integrated Multi-tasks for 
Image-guided Surgical Automation in Laparoscopic Hysterectomy Dataset (AutoLaparo)}
The \href{https://autolaparo.github.io/}{AutoLaparo} dataset \citep{autolaparo} supports image-guided surgical automation in laparoscopic hysterectomy, offering three sub-datasets: phase recognition, laparoscope motion prediction, and instrument segmentation with key anatomy annotation. The 21 procedures are recorded at 25 fps with a resolution of 1920x1080 pixels. These procedures are annotated with phase labels (7 phases). The laparoscope motion prediction sub-dataset consists of 300 clips, extracted from phases 2-4 of recorded procedures, each annotated with one of seven motion modes (up, down, left, right, zoom-in and zoom-out). The segmentation sub-dataset includes instruments (4 instruments) and key anatomy (1 anatomy) segmentation for keyframes in the motion prediction clips.

\subsection{Augmented Reality Tool Network dataset (ART-Net)}
The \href{https://github.com/kamruleee51/ART-Net}{ART-Net} dataset \citep{art_net} is tailored for non-robotic laparoscopic hysterectomy, emphasizing 3D graphics applications. Annotations cover tool presence detection, binary tool segmentation, and 2D pose estimation. Extracted from 29 procedures, the dataset provides frames with and without instruments, annotating these frames for tool presence. Keyframes are annotated for binary tool segmentation and 2D pose in the form of geometric primitives (tool-tip, midline and instrument shaft heatmaps).

\subsection{HeiChole Surgical Workflow Analysis and Full Scene Segmentation dataset (HeiSURF)}
The \href{https://www.synapse.org/#!Synapse:syn25101790/wiki/}{HeiSURF} dataset \citep{heisurf} for day-to-day laparoscopic cholecystectomy includes annotations for phase recognition, full scene segmentation, action recognition, instrument presence detection, and skill assessment. Each procedure in the dataset is annotated for phase (7 phases), and keyframes are annotated for full scene segmentation and actions. Three videos per procedure are extracted and annotated with 5 skill annotations. The HeiSURF dataset is a parent dataset of an earlier dataset - the HeiChole dataset (This did not contain segmentation information) \citep{WAGNER2023102770}.

\subsection{Phase, Step, Instrument, and Atomic Visual Action recognition dataset (PSI-AVA)}
The \href{https://github.com/BCV-Uniandes/TAPIR}{PSI-AVA} dataset \citep{holistic_scene_understanding} focuses on robot-assisted radial prostatectomy for phase recognition, step recognition, instrument presence, instrument bounding boxes, and action recognition. The whole dataset is labelled for phase (11 phases) and step (21 steps) recognition. Keyframes annotated with bounding boxes provide detailed instrument detection (7 instruments) and corresponding actions (16 actions).

\subsection{The Heidelberg Colorectal Dataset for Surgical Data Science in the Sensor Operating Room dataset (HeiCo)}
The \href{https://www.synapse.org/#!Synapse:syn21903917/wiki/601992}{HeiCo} \citep{heico} dataset features colorectal procedures for instrument segmentation and phase recognition. The dataset, derived from 30 procedures, covers proctocolectomy, rectal resection, and sigmoid resection. Phase recognition (14 phases) annotations are provided for the whole dataset. Keyframes with instance segmentation labels are also provided. A 10-second video snippet preceding each annotated keyframe is provided for temporal context. HeiCo contains data on surgical workflow analysis from the sensorOR challenge \citep{heico} and data from the Robust-MIS segmentation dataset \citep{robustmis2019}.

\subsection{The John Hopkins University - Intuitive Surgical Inc. Gesture and Skill Assessment Working Set dataset (JIGSAWS)}
\href{https://cirl.lcsr.jhu.edu/research/hmm/datasets/jigsaws_release/}{JIGSAWS} dataset \citep{jigsaws} contains annotations for manipulator gestures and surgical skills collected from recorded trials of eight medical doctors with varying skill levels performing suturing, knot-tying, and needle-passing task with the da Vinci Robot Surgical System (DSS). The dataset includes kinematic data (76D vector capturing the position, orientation, velocity and angle of the manipulator and camera), synchronized stereo video recordings (each at 30fps for approximately 2 minutes), and annotations for automatic gesture recognition (15 gestures) and skill assessment (a modified form of OSATS \citep{osats}).

\subsection{The MIcro-Surgical Anastomose Workflow recognition on training sessions dataset (MISAW)}
\href{https://www.synapse.org/#!Synapse:syn21776936/wiki/601700}{MISAW} dataset \citep{huaulme2021micro} focuses on multi-granularity surgical activity recognition in micro-surgical anastomosis. The dataset consists of 27 sequences recorded using a stereo-microscope, along with annotations for phase (2 phases), step (6 steps), action (17 actions), instrument presence (1 instrument), and instrument tissue targets (9 targets) for each frame. Synchronized kinematic data are also provided.

\subsection{PEg TRAnsfer Workflow recognition by different modalities dataset (PETRAW)}
The \href{https://www.synapse.org/#!Synapse:syn25147789/wiki/}{PETRAW} dataset \citep{huaulme2022peg} is designed for workflow recognition in peg transfer training sessions. The dataset comprises 150 sequences of peg transfer sessions recorded on a virtual reality simulator, kinematic data, videos, and annotations for semantic segmentation (2 targets and 1 instrument), phase (2 phases), step (12 steps), and action annotations (6 actions).

\subsection{Surgical Instrumentation Segmentation and Action Recognition on Robot-Assisted Radical Prostatectomy dataset (SAR-RARP50)}
The \href{https://www.synapse.org/#!Synapse:syn27618412/wiki/616881}{SAR-RARP50} dataset \citep{psychogyios2023sar} is designed to address data-scarcity challenges in surgical action recognition and tool segmentation for in vivo Robotic Assisted Radical Prostatectomy. The dataset comprises of 50 suturing segments acquired with a DaVinci Si Robot that features a stereo endoscope. The dataset is annotated for instrument segmentation (9 instruments) at keyframes and action (8 actions) for the whole dataset.

\subsection{The CholecT50 dataset}
CholecT50 \citep{NWOYE2022102433_rendevous} is a dataset for furthering the research of tool-tissue interactions, formalized as action triplets \textlangle{}instrument, action, target\textrangle{} for laparoscopic cholecystectomy procedures. 
It consists of 50 laparoscopic cholecystectomy procedures annotated with action triplets. As it is a subset of the Cholec120 dataset, it also contains relevant annotations from Cholec120 such as phase recognition, and tool presence labels for the relevant procedures. CholecT50 is the most comprehensive iteration of action triplet datasets released by the CAMMA research group which also includes  CholecT40 \citep{1nywoye_rec_action_triplets}, and \href{https://github.com/CAMMA-public/cholect45}{CholecT45}. \citet{nwoye2022data} gives in-depth information about the CholecT50 and related datasets, the benchmarks, metrics, and other relevant information.

\subsection{Multi-centric Multi-activity Dataset laparoscopic Roux-en-Y gastric bypass (LRYGB) dataset(MultiBypass140)}
The \href{https://github.com/CAMMA-public/MultiBypass140}{MultiBypass140} dataset \citep{Lavanchy2024} is designed for phase and step recognition, particularly focusing on datasets collected from different hospitals (multi-centric) and the importance of multi-centric datasets for generalization. 
The dataset comprises 140 sequences of laparoscopic Roux-en-Y gastric bypass (LRYGB) surgeries performed at two medical centres and annotations for phase (12 phases), and steps (46 steps). MultiBypass140 is an extension of the Bypass40 \cite{Ramesh2021_phase_step_gastric_bypass} dataset.

\subsection{The Endoscapes Dataset}
The \href{https://github.com/CAMMA-public/Endoscapes}{Endoscapes} dataset was designed for advancing research in automated assessment of surgical scenes, particularly focusing on the Critical View of Safety (CVS), segmentation and object in Laparoscopic Cholecystectomy surgeries \citep{murali2023endoscapes}. 
The dataset comprises 201 videos and annotations for CVS(3 labels), bounding boxes for anatomy(5 classes) and a surgical instrument (1 class), as well as segmentation masks for these bounding boxes (6 classes).

\subsection{SARAS challenge on Multi-domain Endoscopic Surgeon Action Detection dataset (SARAS-MESAD)}
\href{https://saras-mesad.grand-challenge.org/Home/}{SARAS-MESAD} dataset \citep{saras_mesad} facilitates surgical activity recognition research and cross-domain learning. It includes MESAD-Real and MESAD-Phantom sub-datasets, offering annotated frames for instrument detection and action recognition for human prostatectomy and phantom surgeries. This dataset builds on the previous \href{https://saras-esad.grand-challenge.org/}{SARAS-ESAD} dataset \citep{bawa2021saras}.

\subsection{The EndoVis-18-VQA dataset} \label{sec:endovis_2018_vqa_dataset}
The \href{https://github.com/lalithjets/SurgicalGPT}{EndoVis-18-VQA} dataset  \citep{seenivasan2022surgical} was designed for advancing research in visual question answering using the Endovis2018 dataset. 
The dataset comprises images from the 14 sequences in the Endovis2018 dataset and contains classification-based questions and answers about tissues, actions and tool locations from the Endovis2018 annotations. This dataset is the more recent version of Endovis2018-with-interactions discussed in \secref{sec::report_generation}

\subsection{Cholec80-VQA}
The \href{https://github.com/lalithjets/SurgicalGPT}{Cholec80-VQA} dataset was designed for advancing research in visual question answering using the Cholec80 dataset \citep{seenivasan2022surgical}. 
The dataset comprises images from 40 sequences in the Cholec80 dataset and contains classification-based questions on surgical phase and instrument presence from the Cholec80 annotations.

\subsection{PSI-AVA-VQA}
The \href{https://github.com/lalithjets/SurgicalGPT}{PSI-AVA-VQA} dataset was designed for advancing research in visual question answering using the PSI-AVA dataset \citep{seenivasan2023surgicalgpt}. 
The dataset comprises images from 8 sequences in the PSI-AVA dataset and contains classification-based questions on surgical phase, step, and location annotations from the PSI-AVA annotations.

\subsection{Surgical Scene Graph-based dataset (SSG-VQA)}
The \href{https://github.com/CAMMA-public/SSG-VQA}{SSG-VQA} dataset was designed for advancing research in visual question answering, in laparoscopic cholecystectomy surgeries \citep{yuan2024advancing}. 
The dataset comprises images from the 45 sequences of the CholecT45 dataset \citep{NWOYE2022102433_rendevous}, interaction labels(\textlangle{}instrument, action, target\textrangle) from the CholecT45 dataset, bounding boxes generated by pseudo-labelling using models trained on m2cai16-tool-locations \citep{jin2018tool} and CholecSeg8k \citep{cholecseg8k}, and questions for various tasks such as questions on spatial locations of instruments and tissues, relationships between objects in a scene,  generated by a well-curated question template engine with strategies to address the class imbalance and remove poorly formulated questions that are not challenging. Questions were designed to be more complex than other surgical VQA datasets available.

\section{Discussion and conclusion} \label{sec::discussions_and_conclusions}
In this section, we consolidate our insights and observations from the MIS-related papers in \secref{sec::mtl_surgical} and connect them with generic MTL techniques for natural images introduced in \secref{sec::multitask_learning_natural_images}.
We also draw parallels with recent trends in the deep learning community.

\subsection{  Learning multiple tasks together in the MIS domain }

A prominent observation from the reviewed papers in \secref{sec::mtl_surgical} is that there are successful applications of MTL in minimally invasive surgeries (MIS). . We have observed several methodologies for learning perceptual tasks together in this context 
\cite{simultaneous_depth_tool_seg,scalable_joint_detection,islam2020apmtl,baby2023forks,zhao2022trasetr,das2023multi,msdesis}. While most reported positive outcomes \cite{simultaneous_depth_tool_seg,scalable_joint_detection,islam2020apmtl,baby2023forks,zhao2022trasetr,das2023multi}, others indicate negative transfer effects or negligible improvements \cite{msdesis}, demonstrating the need for in-depth investigations on which tasks assist each other or regularly cause negative transfer, similar to the work of \citet{standley2020tasks}.  Furthermore, we found a lack of studies focusing on learning perceptual tasks like motion flow, and normal estimation, suggesting an unexplored research direction or that these tasks may not work well in the multitask learning framework. 

\begin{figure}[htb!]
   \centering
  \includegraphics[width=0.48\textwidth]{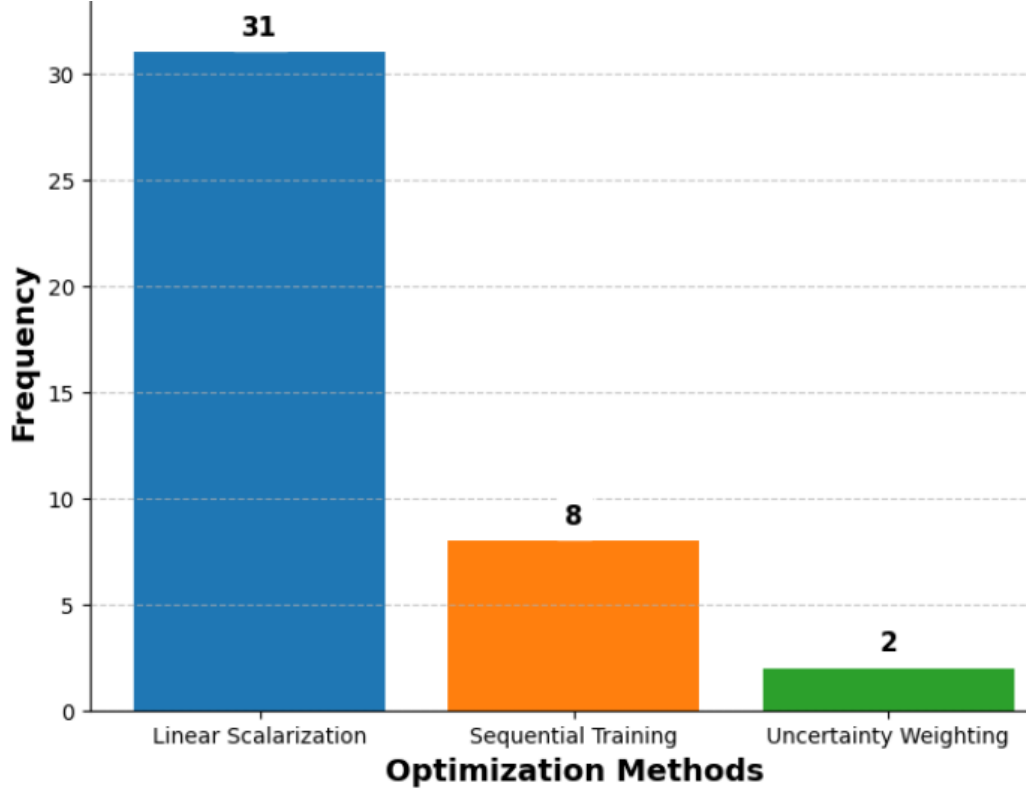}
  \caption{ Prevalence of linear scalarization, sequential training, and uncertainty weighting optimization approaches among reviewed surgical vision works.} 
  \label{fig:optimization_frequency}
\end{figure}

The most common optimization method which was utilized in the reviewed surgical vision works was linear scalarization. Custom sequential training schemes were also utilized, followed by the uncertainty weighting approach \cite{uncertainty_weighting}. \figref{fig:optimization_frequency} illustrates the prevalence of these three methods in the reviewed literature. Hence, an interesting future research question would be to systematically compare the efficiency of various multitask optimization techniques for learning multiple tasks in MIS compared to standard linear scalarization with grid search similar to the works of \citet{linear_scalarization_defense} and \citet{xin2022current}. There is a noted rarity of multitask optimization methods in MIS, which should be investigated.

The architecture of how multiple tasks are learned together in the context of MIS presents an interesting observation. In all surgical vision MIS papers reviewed, the standard approach involves the use of hard parameter sharing.  The architectures used in these surgical vision works can be categorized into four main types: `shared encoder with multiple task branches',` multistage network', `shared encoder with interacting task branches', and `unique'. The `unique' category includes architectures with distinctive configurations, such as multiple encoders or transformers, where the task structure deviates significantly from conventional designs.  Details on these architecture types for each reviewed work can be found in \tabref{perceptual_tasks:table}, \tabref{tracking_and_control:table}, \tabref{workflow:table}, \tabref{workflow_optimization_tasks:table}, \tabref{skills_tasks:table} and \tabref{report:table}.

\begin{figure}[htb!]
   \centering
  \includegraphics[width=0.48\textwidth]{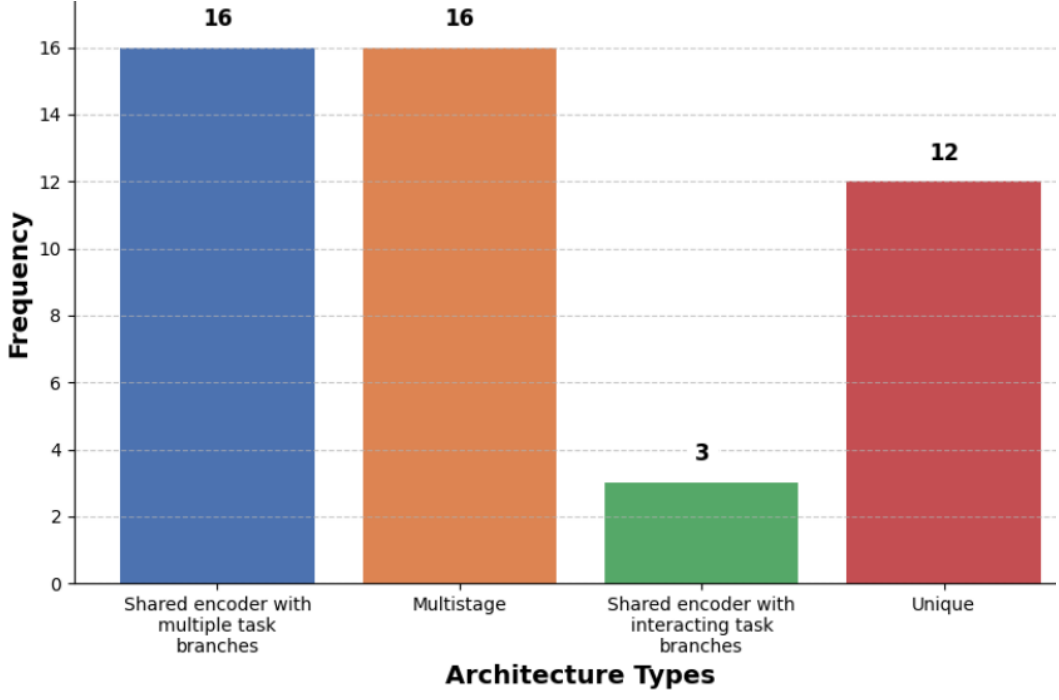}
  \caption{Prevalence of the architecture types in the reviewed surgical vision works.  } 
  \label{fig:architecture_frequency}
\end{figure}

\figref{fig:architecture_frequency} illustrates the prevalence of each architecture type among the reviewed works, with `shared encoder with multiple tasks branches' and `multistage' architectures as the most common architectures.  However, one noticeable absence is the use of soft parameter sharing in MTL for MIS. Soft parameter sharing can allow tasks to share information at different levels and determine how to share information as discussed in  \secref{sec::multitask_learning_natural_images}. This could be because of the complicated nature of soft parameter sharing networks or that hard parameter sharing is just good enough in most MIS scenarios. An area of exploration in future research could be to determine if soft parameter sharing can provide advantages over hard parameter sharing in the MIS context by comparing and contrasting the performance of both architectural styles on minimally invasive surgeries.

Another observation pertains to the impact of auxiliary tasks on learning accuracy. While many studies have shown the advantages of auxiliary tasks in enhancing task performance \cite{lin2020lc,multi_HOG,ss_depth_estimation_laparo}, it is essential to remain critical and consider possible negative transfers. The choice between directly incorporating domain-specific information into the primary network or predicting this information as an auxiliary task should be further explored.  For instance, using contour prediction as an auxiliary task  \cite{multi_contour} or applying boundary loss \cite{kervadec2019boundary} may yield different outcomes in terms of both efficiency and informativeness. 

Another noteworthy point is that only a few works among the reviewed surgical vision papers incorporate information flow between decoders in the MIS field, specifically through the `shared encoder with interacting task branches' architecture \cite{NWOYE2022102433_rendevous,sharma2022RendezvousInTime,seenivasan2022global}. This trend is illustrated in \figref{fig:architecture_frequency}
, which displays the prevalence of different architectures among the reviewed surgical vision works. The additional connectivity between decoders can facilitate inter-task relationship learning, which is a unique advantage of MTL. By allowing tasks to influence and inform each other, models can develop a deeper understanding of the underlying relationships between tasks and potentially improve overall performance. While learning better representations through MTL is valuable, it is essential to acknowledge that there are other learning paradigms, such as self-supervised learning, which excel in representation learning. However, the specific advantages of MTL in MIS, including inter-task relationship learning, make it a compelling choice for solving complex surgical tasks.

\subsection{Multitask learning for automatic camera control}
The discussion on automatic camera control in the context of MIS raises some interesting points. While the scanpath prediction task has been a significant development, it may not provide a comprehensive solution for automating camera movements in these surgeries. The approach of focusing primarily on surgical instruments \citep{learning_look_tracking,Islam2020_ST_MTL}, while valuable, may overlook the importance of capturing the surgical field comprehensively, including tissues and organs. Surgical procedures often involve dynamic movements where instruments enter and exit the field of view, and the camera's focus may need to adapt to these changes, as detailed by \citet{prostatectomy_video}. This highlights the complexity of the task, where understanding the surgical context and deciding what to focus on is crucial.

An emerging approach to automating camera control is camera imitation and surgical intent learning \citep{10.1007/978-3-031-43996-4_21,9772402}. While this field is still relatively new, and there are only a handful of papers exploring it, it holds promise for improving camera control in minimally invasive surgeries. The release of datasets like AutoLaparo is expected to drive more interest and research in this area. Incorporating MTL into camera control, where tasks include predicting future segmentation and motion for the camera, is great. The future development of this field may involve adding more tasks, such as predicting ongoing and future actions, surgical steps and phases, to provide a holistic solution for camera control.

As the field of automatic camera control in MIS matures, it will be fascinating to see how researchers tackle the challenges of understanding and target domain generalization of surgical scenes to enhance camera movements and, ultimately, improve the surgical experience and outcomes.

\subsection{Multitask learning for  surgical video workflow analysis }

The evolution of  surgical video workflow analysis from phase detection to action triplets and multi-granular activity detection is a notable development. This progression underscores the importance of understanding the intricate relationships between different aspects of surgical activities, leading to improved recognition of complex surgical procedures and their contextual understanding. \tabref{evolution_activity:table} presents the reviewed studies in surgical video workflow analysis (along with the publication year), dataset used, tasks addressed, and architectural choice.

\begin{table}[htb!]
\caption{Evolution of Surgical Activity Recognition.} \label{evolution_activity:table}
\resizebox{\columnwidth}{!}{
\begin{tabular}{|l|l|l|l|}
\hline 
 Study &  Tasks & Dataset & Architecture \\
\hline
\citet{twinanda2016lstm} & \thead[l]{phase recognition\\tool presence detection} & Cholec80 & multistage  \\
\hline 
\citet{EndoNet} & \thead[l]{phase recognition\\tool presence detection}  & Cholec80  & multistage  \\
\hline 
\citet{multitask_connectionism}& \thead[l]{phase recognition\\tool presence detection}  & Cholec80  & multistage   \\
\hline 
\citet{tecno} & \thead[l]{phase recognition\\tool presence detection} & Cholec80 &  multistage \\
\hline 
\citet{multi_task_recurrent_conv}  & \thead[l]{phase recognition\\tool presence detection} & Cholec80 &  multistage  \\
\hline
\citet{Sanchez-Matilla2022_data_centric}  & \thead[l]{phase recognition\\scene segmentation} & \thead[l]{Cholec80\\CholecSeg8k} &  multistage  \\
\hline

\citet{1nywoye_rec_action_triplets}  & surgical action triplet recognition &  CholecT40 &  encoder with multiple branches  \\
\hline 
\citet{NWOYE2022102433_rendevous}  & surgical action triplet recognition & CholecT50 &  shared encoder with interacting task branches  \\
\hline 
\citet{sharma2022RendezvousInTime}  & surgical action triplet recognition & CholecT45 &  shared encoder with interacting task branches  \\
\hline 
\citet{multi_self_distillation}  & surgical action triplet recognition & CholecT45 &   encoder with multiple branches  \\
\hline 
\citet{sharma2023surgical}  & surgical action triplet detection & CholecT50 &  multistage  \\
\hline 

\hline 
\citet{holistic_scene_understanding}  & \thead[l]{phase recognition \\ step recognition \\ Action recognition \\ Instrument detection} & PSI-AVA &  multiple encoders with multiple task branches  \\
\hline 

\end{tabular}
}
\end{table}

In the early stages, much of the focus was on phase detection, which was partially influenced by the availability of datasets like Cholec80. Early models were primarily two-stage systems, incorporating temporal modelling to capture the dynamics of surgical activities. A significant shift has been observed in recent research, where surgical activity detection has been framed as the prediction of surgical action triplets, which include instrument, verb, and target. This approach provides a more detailed and informative way to describe surgical activities. It also highlights the need for understanding complex interactions between these components during surgical procedures.

The exploration of multi-granularity research, as exemplified by the PSI-AVA and MISAW datasets, offers valuable insights into recognizing different levels of surgical activities. It would be interesting to see how these methods that are surgery-specific and models that are trained specifically for multiple granularities of this specific procedure are generalized to other surgical procedures, surgeon styles, and surgical contexts.  

As the field of surgical activity recognition continues to evolve, addressing the challenge of adapting models to different surgeries and achieving broader generalization is essential. Future research may focus on creating more versatile models that can understand and recognize surgical activities across various surgical procedures, improving the practical utility of these techniques in clinical settings.

\subsection{Multitask learning for report generation}
The application of MTL to surgical report generation is a promising avenue for improving the documentation and record-keeping aspects of minimally invasive surgeries. This complex task requires the generation of detailed and organized information about the events and procedures that occur in the intra- or pre-operative period.

Two approaches have been explored as discussed in \secref{sec::report_generation} : one involving training scene graphs with segmentation models to obtain reports \citep{seenivasan2022global}  and the other incorporating scene graphs and frame captioning techniques to generate reports \citep{mobarakol_task_aware}. These approaches are initial steps towards addressing the challenge of surgical report generation, and they have shown potential for producing structured information from surgical data.

However, a more sophisticated dataset that captures the intricacies of real-life surgical reports and allows for a transition from frame-to-frame inferences to full video inferences is essential for advancing this field, as the currently used dataset, EndoVis2018-with-interactions, is limited by its reliance solely on pixel-level information from the images in EndoVis2018, as described in \secref{sec:endovis_2018_vqa_dataset}

\subsection{Datasets, Unification and Ethics.}
For multitask problems like surgical action triplets, there exists a standard benchmark dataset—the CholecT50. Designed for solving multiple tasks together, CholecT50 is large, encompassing over 100k frames across 50 surgeries, and includes metrics for evaluating both single tasks and relationships between tasks. It also provides a  \href{https://github.com/CAMMA-public/ivtmetrics}{metrics library} , facilitating easy comparison between methods and lowering entry barriers. The CholecT50 dataset is part of the Cholec series, which features similar annotation styles, shared images, and metrics, making it easily extendable.

As large models that solve multiple tasks become more prevalent in the field, the creation of more extensive multitask datasets will be essential. We encourage collaboration within the community to build comprehensive, unified datasets, benchmarks, and metrics. This collaborative effort will enhance the robustness and comparability of research outcomes.

Creating larger datasets involves challenges such as data collection, privacy, patient safety, and regulatory concerns.  Collaboration with surgeons, institutions, regulatory bodies and the general public is crucial to address these issues. More initiatives that inform the public about data usage, its potential to improve healthcare, and the associated ethical concerns, and solutions need to be raised. Addressing regulatory concerns involves ensuring compliance with data protection laws and ethical guidelines, which can be facilitated through robust anonymization processes and transparent collaboration with regulatory bodies to establish clear protocols for data usage and sharing.  Programs like the \href{https://www.cvschallenge.org/data-donation-video}{SAGES video donation program}, which promotes the donation and anonymization of surgical videos and fosters collaboration among researchers, surgeons, and hospitals, are recommended.

\subsection{Large models in surgical scene understanding}
The trend towards large models with the capacity to solve multiple problems by predicting a universal task is an intriguing development in the field of computer vision. Language models like the GPT series \citep{radford2019language}  have demonstrated the potential of understanding the nature of language through a single pretext task (next-token prediction) and subsequently applying that understanding to various language-related tasks. This notion of having a universal pretext task that can be leveraged for solving diverse problems is both promising and efficient.

While the concept of universal pretext tasks has gained traction in natural language processing, it is noteworthy that the vision domain is yet to have a similarly unified and versatile framework. Recent efforts, such as the SAM \citep{kirillov2023segment} and models derived from SAM, bridge this gap for segmentation, but this is just a single visual task. Looking forward, it is possible that developing a universal vision task and devising efficient methods for querying this universal task for specific applications could become a foundational approach for visual models. This paradigm would allow for a more streamlined and versatile way of addressing multiple vision-related challenges and potentially lead to breakthroughs in the field and, by extension, the computer vision for the MIS.

The generalizability of large models for multiple tasks is great. However, a main issue with large models in the surgical scenario is that for the networks to be utilized in Operating rooms (ORs), they need to perform inference in real-time and fit into devices and computers in the OR, which would be a cost-inefficient solution.

\subsection{Real-time deployment}
The deployment of artificial intelligence (AI) in surgery is in its nascent stages compared to fields like radiology \citep{maier2022surgical}. Despite the potential benefits, there have been relatively few clinical trials exploring AI for intraoperative assistance \citep{varghese2024artificial,auloge2020augmented,mascagni2024early}.

The primary challenges hindering the widespread adoption of AI in surgery are multifaceted. Technically, collecting and managing large volumes of surgical data, ensuring real-time processing, and achieving low-latency inference are significant hurdles. Integrating these AI systems into existing surgical workflows without disruption is also a considerable challenge \citep{auloge2020augmented}. Moreover, regulatory and ethical challenges, including ensuring compliance with stringent medical standards, addressing ethical concerns, and maintaining data privacy, require careful consideration. Operational challenges, such as training surgical staff, managing costs, and upgrading infrastructure, further complicate the deployment of AI in surgery.

Multitask learning (MTL) presents a promising solution to some of these challenges, particularly in addressing the need for low-latency inference. Several MTL models reviewed in this survey report speeds exceeding 20 frames per second for multiple tasks while also reducing the number of parameters required by sharing encoders, as shown in \tabref{perceptual_tasks:table}, \tabref{tracking_and_control:table}, \tabref{workflow:table}, \tabref{workflow_optimization_tasks:table}, \tabref{skills_tasks:table}, \tabref{report:table}, and \tabref{large_models:table}.However, more work is required to fully realize the potential of MTL in real-world surgical applications. Unification in datasets and metrics, continued research, development, and collaboration are necessary to address these challenges and leverage AI and MTL to enhance surgical outcomes and efficiency.

\subsection{Conclusion}
In conclusion, MTL has firmly established itself as an important paradigm within the domain of minimally invasive surgeries. Its influence extends across various facets of this specialized field. 

This review analysed the applications of the MTL paradigm to minimally invasive surgeries. Firstly, the review gave an introduction to MTL and its objectives. Secondly, a detailed exploration of six distinct areas where MTL is applied in MIS was provided. Thirdly, the datasets that support MTL for MIS were presented. Lastly, we discussed some of the inferences and interesting observations on MTL and minimally invasive surgeries.

\section*{Acknowledgments}
\paragraph{Declaration of Interests}
TV is a co-founder and shareholder of Hypervision Surgical Ltd, London, UK.
The authors declare that they have no other conflict of interest.

\paragraph{Ethics approval} For this type of study, formal consent is not required.

\paragraph{Informed consent} This article does not contain patient data.

\paragraph{Funding Sources} This work was supported by core funding from the Wellcome/EPSRC [WT203148/Z/16/Z; NS/A000049/1] and, Tongji Fundamental Research Funds for the Central Universities. OA is supported by the EPSRC CDT
[EP/S022104/1].
TV is supported by a Medtronic / RAEng Research Chair [RCSRF1819$\backslash$7$\backslash$34].

\paragraph{Open Access} For the purpose of open access, the authors have applied a CC BY public copyright licence to any Author Accepted Manuscript version arising from this submission.

\bibliographystyle{model2-names.bst}\biboptions{authoryear}
\bibliography{refs}

\end{document}